\documentclass[12pt]{article}

\usepackage{newtxtext,newtxmath}

\usepackage{graphicx}

\usepackage[letterpaper,margin=1in]{geometry}

\renewenvironment{abstract}
	{\quotation}
	{\endquotation}

\date{}

\makeatletter
\renewcommand{\fnum@figure}{\textbf{Figure \thefigure}}
\renewcommand{\fnum@table}{\textbf{Table \thetable}}
\makeatother

\usepackage{scicite}

\usepackage{url}

\def\scititle{
Three-dimensional hydro-cluttered locomotion \\by an undulatory robot
}
\title{\bfseries \boldmath \scititle}

\author{
Tianyu Wang$^{1,2,3,\dagger}$,
Matthew Fernandez$^{3,\dagger}$,
Galen Tunnicliffe$^{1,3}$, \and
Nikolas Cornell$^{3}$,
Justin Duong$^{4}$,
Donoven Dortilus$^{1,3}$,\and
Zhaochen J. Xu$^{2}$,
Patricia Meza$^{5}$,
Sean Lublinsky$^{2}$,
Darsh Parikh$^{4}$,\and
Jianfeng Lin$^{2}$,
Emily Grace$^{6}$,
Daniel I. Goldman$^{1, 2, \ast}$\and
\small$^{1}$Institute for Robotics and Intelligent Machines, Georgia Institute of Technology, Atlanta, GA 30332, USA.\and
\small$^{2}$School of Physics, Georgia Institute of Technology, Atlanta, GA 30332, USA.\and
\small$^{3}$George W. Woodruff School of Mechanical Engineering, Georgia Institute of Technology, Atlanta, GA 30332, USA.\and
\small$^{4}$School of Electrical and Computer Engineering, Georgia Institute of Technology, Atlanta, GA 30332, USA.\and
\small$^{5}$Department of Mechanical and Industrial Engineering, Northeastern University, Boston, MA 02115, USA.\and
\small$^{6}$Ransom Everglades School, Coconut Grove, FL 33133, USA.\and
\small$^\ast$Corresponding author. Email: daniel.goldman@physics.gatech.edu\and
\small$^\dagger$These authors contributed equally to this work.
}

\begin{document} 

\maketitle

\begin{abstract} \bfseries \boldmath
Aquatic robots have expanded human access to underwater environments, yet many underwater spaces contain obstacles that can disrupt open-water locomotion. In ``hydro-cluttered" environments, water is interspersed with rigid and flexible clutter, making body-obstacle contact unavoidable. Operating in these spaces requires robots that can regulate and exploit contact, but this regime remains difficult to model or simulate. Building on recent advances in mechanical intelligence in terradynamically capable limbless robotics, we develop principles for 3D aquatic locomotion using AquaMILR, an elongate limbless robot that combines bilateral cable-driven actuation, programmable body compliance, distributed depth regulation, corrosion-resistant enclosures, and onboard power and electronics for untethered field operation. Systematic robophysical experiments reveal that programmable body compliance regulates body deformation and converts body-environment interactions into fast, robust, forward progression across increasing hydro-clutter constraint strength. Depth regulation provides three-dimensional access, allowing the robot to bypass clutter, recover from obstruction, and continue through otherwise inaccessible routes. In potential jamming scenarios, emergent inertia-induced rolling acts as a spontaneous recovery mechanism, freeing the robot from clutter that would otherwise lead to failure and allowing locomotion to continue without additional control. Tests of the robot in an aquatic mangrove field demonstrate that these principles transfer to practical operation, enabling navigation and onboard visual inspection of inaccessible root zones. These results establish principles for hydro-cluttered locomotion and a design paradigm in which aquatic robots exploit environmental complexity as a locomotor resource.

\end{abstract}

\section*{INTRODUCTION}

Aquatic robots enable exploration, inspection, and intervention in places that are difficult or dangerous for humans to reach~\cite{zereik2018futuretrends,fu2026challenges,nauert2023inspection}. Such devices have advanced through a diversity of designs, including fish-like swimmers for fast propulsion~\cite{barrett1999drag,marchese2014autonomous,katzschmann2018sofi,zhu2019tunabot,zhong2021tunable,White2021tunabot3,aragaki2023biomimetic,du2015robot}, propeller-driven AUVs and ROVs for long-duration inspection~\cite{mccammon2026autonomous,yoerger2021hybrid,patel2025comprehensive,cepeda2023offshoreStructureInspections,Zhang2021DeepwaterDamInspection,teague2018mineral,Lawson2023AerialAquaticAcoustic}, deep-sea robots for extreme environment exploration~\cite{li2021self,pan2025miniature,li2025plasticized,wang2026marianaTrench, Liang2022DeepSeaEquipment,Li2023DeepSeaSoftRobots}, multimodal robots for cross-medium locomotion~\cite{Baines2022AdaptiveMorphogenesis,xia2025toward,crespi2008online,spino2024towards,zufferey2022between,debruyn2020medusa,chen2017biologically}, and systems for compliant actuation and dexterous manipulation~\cite{qu2024recent,Li2026BioinspiredUnderwaterSoft,zhang2023bioengineering,wang2023versatile,Lu2025variableStiffness,khatib2016ocean,stuart2017ocean}. Such robots are predominantly designed and tested in open-water conditions where hydrodynamic interactions dominate, and years of theory and design~\cite{watson2020localisation,raj2016fish,Qiao2026AUVSonarReview,fan2019robotic,ren2021research} have made possible the discovery of principles governing their locomotion.

\begin{figure}[t!]
\centering
\includegraphics[width=1\textwidth]{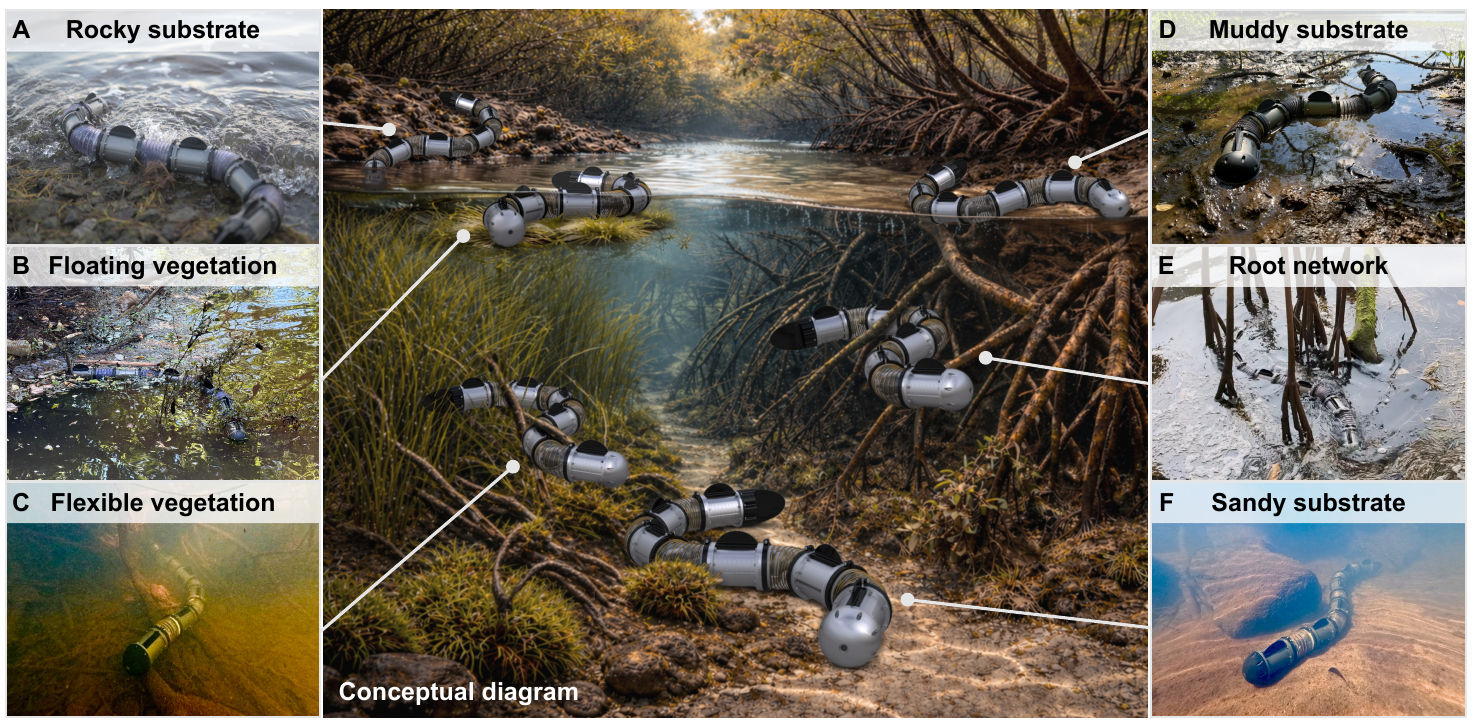} 	
\caption{\textbf{Operating profile of AquaMILR in hydro-cluttered aquatic habitats.}
AquaMILR targets confined, boundary-rich, and hard-to-access aquatic environments where swimming in water is coupled to body interactions with heterogeneous substrates, vegetation, and rigid clutter. The central conceptual diagram illustrates a representative nearshore hydro-cluttered habitat containing multiple mechanically distinct aquatic interaction regimes. The surrounding insets are field images of AquaMILR operating in real aquatic environments: (\textbf{A}) rocky substrate, (\textbf{B}) floating vegetation, (\textbf{C}) flexible vegetation, (\textbf{D}) muddy substrate, (\textbf{E}) root network, and (\textbf{F}) sandy substrate.}
\label{fig:fig1}
\end{figure}

However, many ecologically and practically important aquatic environments are more complex than the open-water settings in which underwater robots are often developed and tested~\cite{wong2018autonomous,mazzini2014experimental,besson2022towards}. From vegetated wetlands, reefs, kelp forests, seagrass beds, and mangroves to aquaculture farms, submerged infrastructure, culverts, shipwrecks, and debris fields, many aquatic environments are ``hydro-cluttered", with water interspersed among branching structures, compliant vegetation, suspended objects, heterogeneous substrates, and confined passages (Figure~\ref{fig:fig1}). In these settings, obstacles are not discrete impediments to be avoided or isolated targets to be manipulated, but part of the locomotor medium through which the robot must move. Contact is unavoidable, persistent, and distributed along the body during locomotion. Robotic operation requires movement through dense structures while repeatedly negotiating contact, deformation, confinement, and post-contact recovery.

The hydro-cluttered regime differs fundamentally from terrestrial and open-water cases. Terrestrial locomotion in cluttered media, broadly studied under the terradynamics framework~\cite{li2013terradynamics}, typically involves frictional contact interactions confined to a quasi-2D surface. Across a diversity of animals and robots, friction dominated dissipative forces dominate such that inertial effects are negligible and coasting is absent~\cite{rieser2024geometric,hu2009mechanics}. Open-water swimming of centimeter to meter scale systems, by contrast, takes place in fluid environments where inertial effects dominate, and propulsion is governed primarily by hydrodynamic forces. The hydro-cluttered regime combines both: it retains the inertial, three-dimensional fluid dynamics of open-water swimming while introducing repeated solid contacts that couple body deformation, environmental constraint, and post-contact recovery. Unlike either case alone, in hydro-cluttered locomotion, the body must absorb and recover from contact without the stabilizing damping of terrestrial environments, while sustaining hydrodynamic propulsion despite contact-induced instabilities in body posture and orientation. Thus, locomotion depends not on hydrodynamic propulsion alone, but on how the body navigates, absorbs, and exploits these interactions, making hydro-cluttered locomotion difficult to predict using computational approaches~\cite{luo2023mantaSFI,tian2020cfd,dowell2001modeling,hou2012numerical,ager2021computationalFSI,bazilevs2013computational,liu2017vortexDynamics,guo2023vortexDynamics2}, especially when obstacles can move, deform, and contact the robot body at multiple points. Hydro-cluttered locomotion thus emerges as a regime that requires systematic study through a robophysical approach~\cite{aguilar2016review,aydin2019physics}, in which effects of robot mechanics and environmental constraint strength can be systematically varied and investigated.

In biological systems, undulation (anguilliform motion) is a common locomotor strategy for hydro-clutter navigation across scales. From meter-scale eels~\cite{tytell2004acceleration}, lampreys~\cite{williams1989locomotion}, and sea snakes~\cite{graham1987surface} to centimeter-scale leeches~\cite{chen2011mechanisms} and lancelet~\cite{stokes1997larval} to millimeter-scale nematodes~\cite{korta2007mechanosensation}, elongate aquatic animals use body undulation~\cite{lauder2005hydrodynamics, gillis1996anguilliformSwimming,pierce2026dispersion,stin2024form} to generate propulsion while conforming to surrounding vegetation, rocks, substrates, and confined passages. Limbless robots modeling these organisms have been developed for a variety of swimming modes \cite{li2023underwater,kelasidi2016innovation,parra2023modular,liu2023nonbiomorphic,yamada2005development,thandiackal2021emergence,liu2026energy,liljeback2014mamba,deng2023development,vogel2008modes}, yet their locomotor performance still falls short of biological counterparts. One feature of elongate animals absent in most such robot designs is bilateral actuation: this muscular architecture plays a key role in tuning body stiffness and mechanical response during locomotion~\cite{tytell2010interactions,tytell2018body,petzold2011caenorhabditis,long1998muscles,sfakiotakis1999review}, suggesting it can be critical for navigating complex, contact-rich environments.

Terrestrial limbless locomotion in cluttered environments has been systematically studied using a bilaterally actuated limbless robot platform~\cite{wang2023mechanical}. This form of actuation models musculoskeletal structures in limbless animals and enables programmable, anisotropic, and quantifiable body compliance, allowing the robot to generate propulsion while passively modulating its response to environmental clutter. At the appropriate compliance level, robot performance matched that of \textit{C. elegans} across a range of obstacle densities. Although \textit{C. elegans} lives in fluid-filled environments, it locomotes in a low Reynolds number (low-coasting) regime~\cite{purcell2014life,rieser2024geometric} where inertial effects are negligible. The robophysical model demonstrated that appropriate body mechanics can handle with environmental complexity in place of sensing and computation---a form of mechanical intelligence that simplifies control in overdamped cluttered terrestrial environments.

We hypothesize that bilateral actuation offers a similar means for inertially dominated (higher Reynolds number) hydro-cluttered locomotion, enabling the robot to regulate and exploit body-environment interactions while maintaining undulatory propulsion. However, transferring this principle from overdamped terrestrial to inertial aquatic environments is nontrivial: locomotion is fully three-dimensional, and inertia, buoyancy, and hydrodynamic forces introduce locomotor dynamics absent in terrestrial clutter. Preliminary attempts to bring bilateral actuation to aquatic limbless robots~\cite{wang2025aquamilr,fernandez2025aquamilr+} demonstrated its potential, but lacked further systematic study of the locomotion principles and the capability to operate in practical field scenarios, leaving the benefits and trade-offs of such a scheme in hydro-cluttered locomotion yet to be identified.

Here we introduce a field-deployable bilaterally actuated limbless robot, AquaMILR (\textbf{Aqua}tic \textbf{M}echanically \textbf{I}ntelligent \textbf{L}imbless \textbf{R}obot), that extends principles of mechanical intelligence to hydro-cluttered locomotion through systematic robophysical study and demonstrates their effectiveness in field deployment (Figure~\ref{fig:fig1}). With this platform, we establish three complementary principles governing locomotion in this regime: programmable body compliance as a quantifiable mechanical variable that can convert environmental contact from an impediment into a source of forward progression, open-loop depth regulation showing that use of the vertical dimension provides modes beyond passive body mechanics for escaping clutter configurations that would otherwise impede locomotion, and emergent inertia-induced body rolling as a spontaneous recovery mechanism under strong clutter constraint. Transferring these principles from laboratory to natural hydro-cluttered environments enables inspection of a submerged mangrove root zone, an environment that can challenge humans and conventional underwater platforms.

\section*{RESULTS}
\subsection*{AquaMILR: An elongate limbless robot for hydro-cluttered locomotion}

\begin{figure}[t!]
\centering
\includegraphics[width=0.7\textwidth]{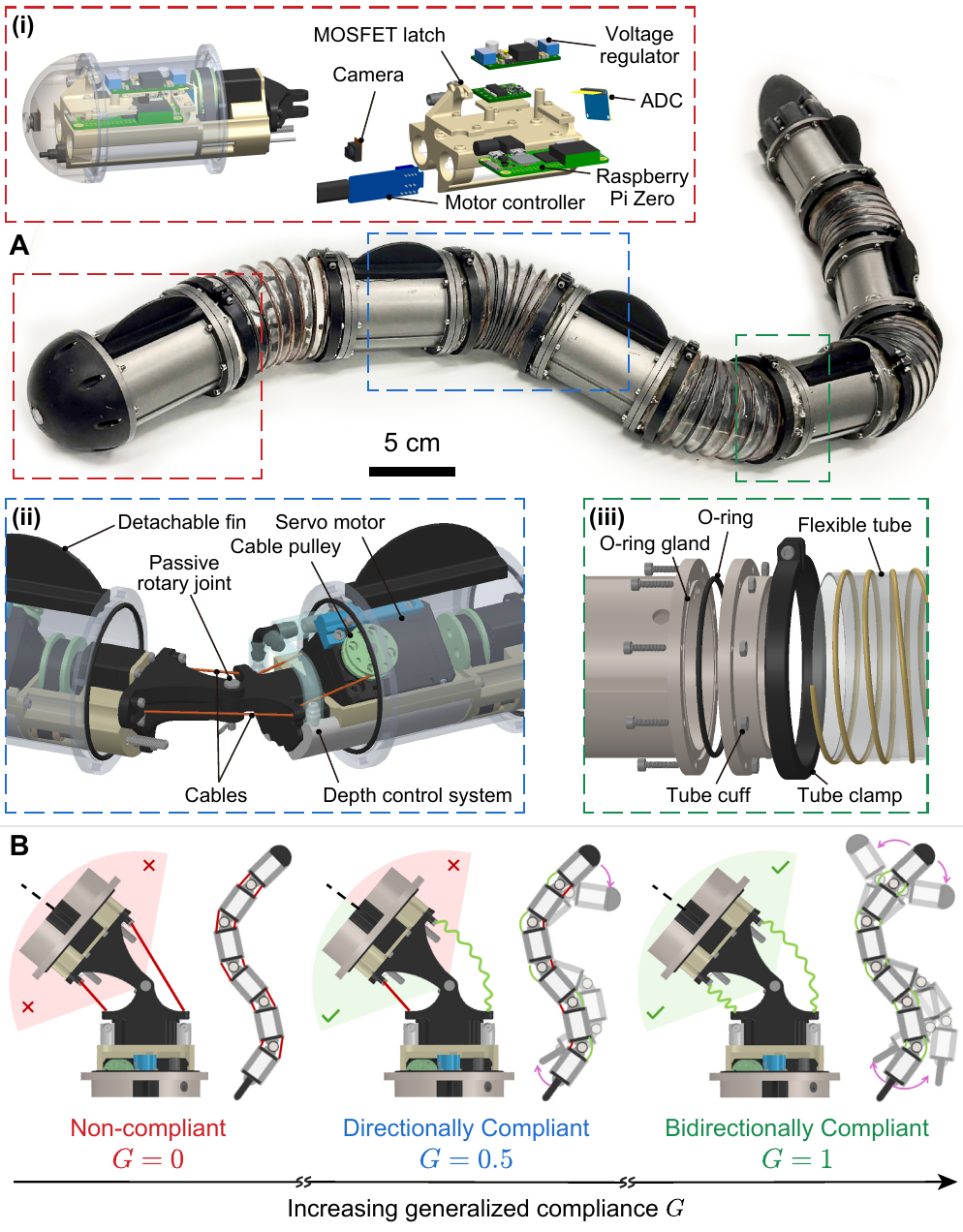} 	
\caption{\textbf{Structural design and programmable compliance of AquaMILR.}
(\textbf{A}) Mechanical architecture of AquaMILR. The assembled robot consists of serially connected body modules. Insets highlight three key structural subsystems: (\textbf{i}) the anterior electronics module, (\textbf{ii}) cable-driven joint modules, and (\textbf{iii}) a waterproof flexible connection that maintains sealing between adjacent modules. (\textbf{B}) Cable-driven joints enable programmable body compliance, quantified by the generalized compliance $G$. In the non-compliant configuration ($G = 0$), passive joint deflection is mechanically constrained. In the directionally compliant configuration ($G = 0.5$), the joints permit passive deflection primarily in the curvature-increasing direction. In the bidirectionally compliant configuration ($G = 1$), the joints permit passive deflection in both directions.}
\label{fig:fig2}
\end{figure}

To investigate hydro-cluttered locomotion, we developed AquaMILR as a modular elongate undulatory limbless robot that integrates bilateral cable-driven actuation, distributed depth regulation, and onboard power and electronics for untethered operation (Figure~\ref{fig:fig2}A). Built on previous terrestrial and aquatic limbless robot prototypes, AquaMILR provides a self-contained experimental platform that supports both systematic robophysical experiments in laboratory and field deployment in natural hydro-clutter environments. A compact body architecture with corrosion-resistant enclosures and self-contained onboard power and electronics supports untethered operation in natural environments, while programmable body compliance and distributed depth regulation modules allow body mechanics and depth to be systematically varied across experiments.

The anterior module contains onboard computation, motor control, power regulation, and a camera for visual inspection during deployment (Figure~\ref{fig:fig2}A-i). Repeated body modules connected in series integrate the cable-driven actuation and depth regulation hardware used throughout the study (Figure~\ref{fig:fig2}A-ii). Flexible sealed connectors maintain watertight sealing through O-ring seals and clamping mechanisms (Figure~\ref{fig:fig2}A-iii). Detailed hardware design and specifications are provided in Materials and Methods.

Cable-driven actuation provides the mechanical basis for whole-body undulation. Each inter-module joint is actuated by a bilateral cable-pulley system with a pair of independently controlled cables, one on each side of the joint. Differential cable motion, in which one cable shortens while the opposing cable lengthens, bends the joint laterally. Coordinated actuation across the serial joints produces traveling waves of body deformation that generate undulatory locomotion.

In each body module, a motor-belt-syringe depth regulation system adjusts module mass through syringe extension and retraction, changing the robot's effective density relative to water (Figure~\ref{fig:fig3}A). With the robot tuned near neutral buoyancy, coordinated syringe movements can shift the robot between rising and sinking during locomotion. This distributed architecture enables depth control during forward locomotion, forming the basis for the three-dimensional navigation experiments described later. In a subset of repeated laboratory experiments, we used a tethered-tail configuration to simplify robot retrieval and avoid repeated battery replacement.

\subsection*{Undulatory motion generation and body compliance programming}

AquaMILR generates locomotion by prescribing a traveling wave of lateral body bending while independently programming the mechanical constraint imposed at each joint. The undulatory gait was defined using a serpenoid shape template~\cite{hirose1993biologically},
\begin{equation}
\alpha_i(t)=A\sin\left(2\pi\xi\frac{i}{N}-2\pi\omega t\right),
\label{eq:serpenoid_aqua}
\end{equation}
where $\alpha_i$ is the suggested angle of the $i$-th joint, $A$ is the commanded joint-angle amplitude, $\xi$ is the spatial frequency, $\omega$ is the temporal undulation frequency, $i$ is the joint index, and $N$ is the number of joints. This feedforward pattern prescribes a periodic retrograde wave along the body and provides the nominal open-loop command used for swimming.

We implemented programmable body compliance through a bilateral cable-driven actuation scheme, using a generalized compliance parameter grounded in comparative biological and robophysical studies of undulatory locomotion~\cite{wang2023mechanical}. Each joint was actuated by independently controlled left and right cables with commanded lengths $L_i^l$ and $L_i^r$. For a suggested angle $\alpha_i$, the corresponding exact cable lengths required to geometrically impose that angle are denoted by $\mathcal{L}_i^l(\alpha_i)$ and $\mathcal{L}_i^r(\alpha_i)$, respectively (see Supplementary Text for the full derivation based on robot geometry). We introduced a joint-level parameter, generalized compliance $G_i \in [0,\infty)$, by selectively lengthening one or both cables relative to these exact lengths as
\begin{equation}
\begin{array}{l}
L_i^l(\alpha_i)=
\left\{
\begin{array}{ll}
\mathcal{L}_i^l(\alpha_i), & \text{if } \alpha_i \leq -\gamma_i, \\[1mm]
\mathcal{L}_i^l[-\min(A,\gamma_i)] 
+ l_0(\gamma_i+\alpha_i), 
& \text{if } \alpha_i > -\gamma_i,
\end{array}
\right. \\[4mm]
L_i^r(\alpha_i)=
\left\{
\begin{array}{ll}
\mathcal{L}_i^r(\alpha_i), & \text{if } \alpha_i \geq \gamma_i, \\[1mm]
\mathcal{L}_i^r[\min(A,\gamma_i)] 
+ l_0(\gamma_i-\alpha_i), 
& \text{if } \alpha_i < \gamma_i,
\end{array}
\right.
\end{array}
\label{eq:gcable_aqua}
\end{equation}
where
\begin{equation}
\gamma_i=(2G_i-1)A.
\label{eq:gamma_aqua}
\end{equation}
The parameter $\gamma_i$ determines which cable is taut or slack at a given 
joint angle: the left cable is taut when $\alpha_i \leq -\gamma_i$ and slack otherwise, while the right cable is taut when $\alpha_i \geq \gamma_i$ and slack otherwise. Thus, $\gamma_i$ sets the angle thresholds beyond which each cable goes slack, directly controlling how much passive joint freedom is admitted around the prescribed wave. $l_0$ is a fixed design parameter that sets the amount of cable slack per unit angular command (see Supplementary Text for more discussion). This formulation separates the prescribed body wave from the passive mechanical freedom available around that wave. The gait parameters $A$, $\xi$, and $\omega$ define the nominal undulatory motion, whereas $G_i$ defines how strongly the $i$-th joint is constrained to follow that motion when perturbed by fluid forces or contact with surrounding structures. Unless otherwise stated, we used a spatially uniform generalized compliance across all joints, such that $G_1=\cdots=G_N=G$.

We used three representative compliance states in the experiments (Figure~\ref{fig:fig2}B), corresponding to qualitatively distinct taut-slack configurations of the cable pair at each joint. At $G=0$, $\gamma=-A$: both cables remain taut throughout the wave cycle, and passive joint deflection is not permitted. In this non-compliant state, AquaMILR reduces to a conventional joint-controlled limbless robot driven to follow the commanded wave. At $G=0.5$, $\gamma=0$: one cable is always taut while the other is slack, depending on the sign of $\alpha_i$. This allows the joint to deflect passively in the curvature-increasing direction while the opposing cable resists deformation in the curvature-decreasing direction, creating an anisotropic mechanical response. At $G=1$, $\gamma=A$: both cables are slack throughout the wave cycle, since $-A \leq \alpha_i \leq A$ by construction and neither taut threshold is reached. The joint is free to deflect passively in either direction under external forcing, permitting the body to deform more freely when in contact with surrounding structures.

\subsection*{Baseline swimming performance across control parameters}

\begin{figure}[t!]
\centering
\includegraphics[width=0.95\textwidth]{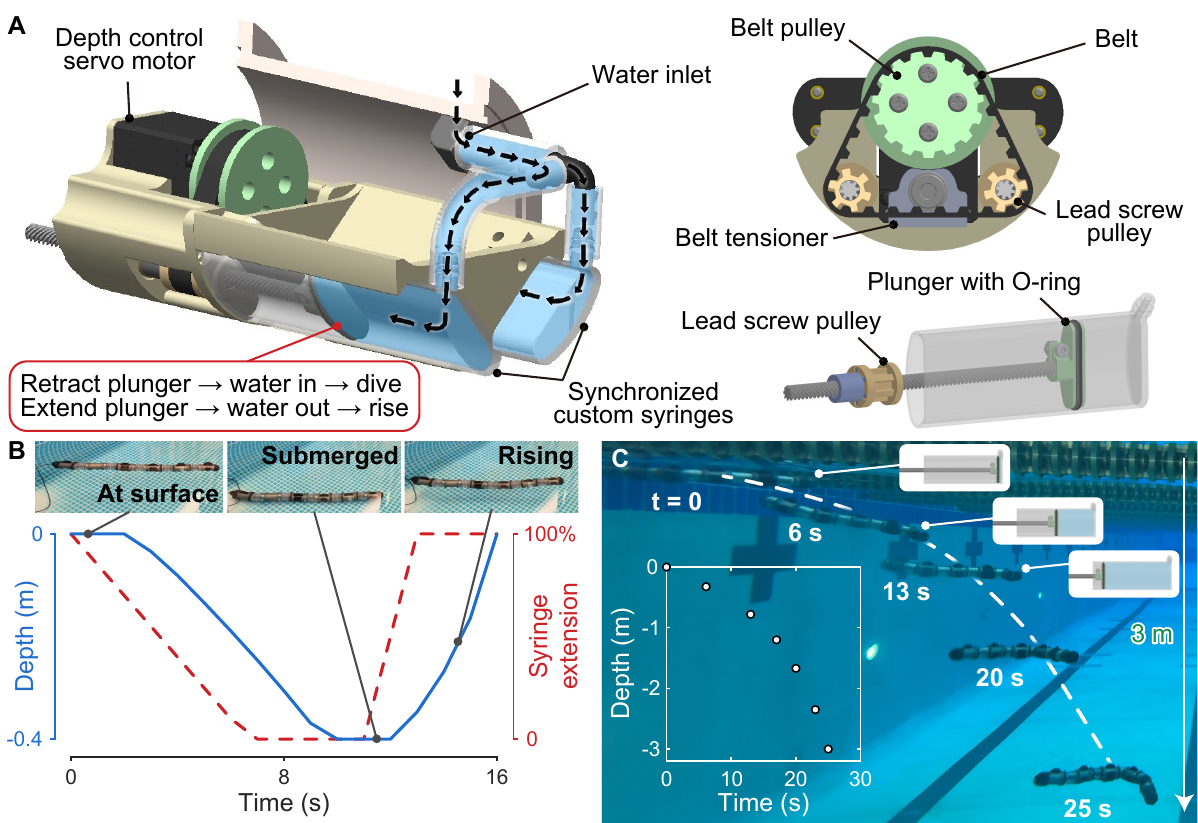}
\caption{\textbf{Distributed syringe actuation enables onboard depth regulation.}
(\textbf{A}) Mechanical design of the depth-regulation module. A depth-control servo motor drives a belt-pulley transmission connected to a lead-screw mechanism, which translates the syringe plungers in synchronized custom syringes. The same mechanism is integrated within the serial body architecture of AquaMILR.
(\textbf{B}) Pool test of vertical motion without body undulation. AquaMILR changes its depth by actuating the onboard syringe modules alone, moving from the water surface to a submerged state and then rising back up to the surface. The blue curve shows the measured vertical position, and the red dashed curve shows the prescribed syringe extension.
(\textbf{C}) Deep-water operational test of the depth-regulation system in a 3-m-deep swimming pool. AquaMILR descends from the water surface toward the maximum depth by drawing water into the onboard syringe modules and generating negative buoyancy. As the robot descends, the system continues to operate under the higher hydrostatic pressure associated with deeper water. The dashed trajectory shows the measured robot path, the inset plots depth as a function of time, and the syringe icons indicate representative plunger states during descent (see also SI Movie 1).}
\label{fig:fig3}
\end{figure}

To access different depths during swimming, each body module contains a depth regulation unit in which a servo motor drives a belt-pulley transmission connected to a lead-screw mechanism, retracting or extending synchronized syringe plungers to draw water in or expel it, shifting the robot between negative and positive buoyancy (Figure~\ref{fig:fig3}A). Coordinated syringe actuation can drive repeated sinking and floating independently of the swimming gait (Figure~\ref{fig:fig3}B, SI Movie 1). With the depth regulation system driving the robot to 3~m depth, AquaMILR maintained coordinated undulatory locomotion throughout the descent despite the elevated hydrostatic pressure, showing that neither the actuation nor the sealing system was compromised at depth. These results confirm that depth regulation and undulatory locomotion operate as independent and mutually compatible systems, with swimming performance remaining fully intact to depths of at least 3 meters (Figure~\ref{fig:fig3}C).

\begin{figure}[t!]
\centering
\includegraphics[width=0.98\textwidth]{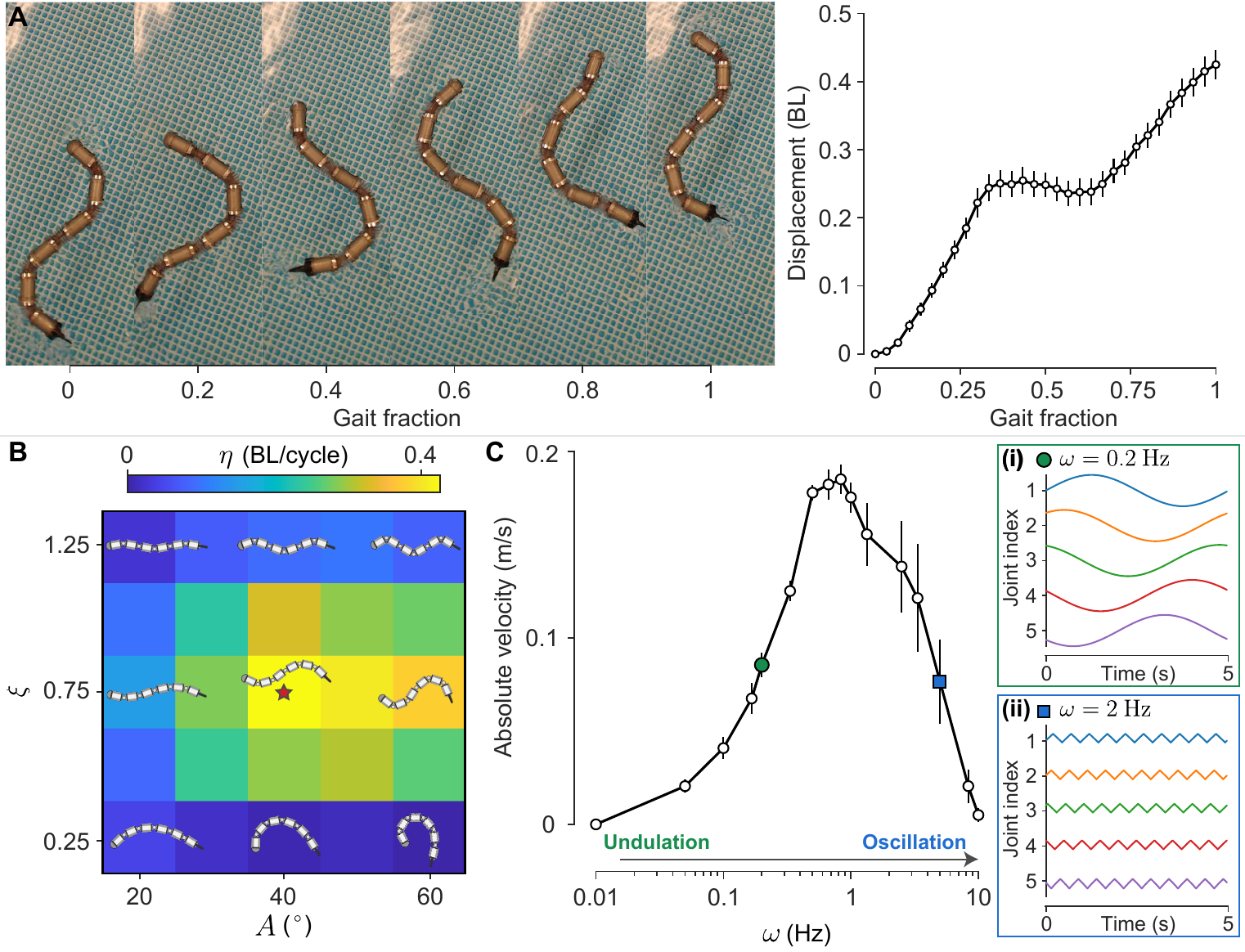} 	
\caption{\textbf{Systematic characterization of open-water swimming performance across body shape parameters and undulation frequency.}
(\textbf{A}) Displacement per gait cycle in open water. The image sequence shows representative robot postures over one cycle, and the plot shows forward displacement normalized by body length. Markers denote mean displacement, and error bars indicate cycle-to-cycle SD over three cycles.
(\textbf{B}) Displacement per cycle as a function of gait amplitude $A$ and spatial frequency $\xi$. Color denotes averaged displacement in body lengths per cycle over five repeated experiments for each combination of parameters, and overlaid robot configurations show representative body shapes. The red star marks the gait used for the frequency sweep in \textbf{C}.
(\textbf{C}) Absolute swimming velocity as a function of undulation frequency $\omega$. Markers denote mean speeds, and error bars indicate SEM. The robot transitions from an undulation locomotor mode at low frequency to an oscillation mode at high frequency. Representative joint angle trajectories at (\textbf{i}) $\omega = 0.2$ Hz and (\textbf{ii}) $\omega = 2$ Hz show frequency-dependent body motion (see also SI Movie 2).}
\label{fig:fig4}
\end{figure}

With depth regulation established as an independent system, we next characterized the baseline swimming performance of the undulatory gait itself. AquaMILR generated stable periodic locomotion under a fixed open-loop gait, with displacement accumulating consistently across cycles (Figure~\ref{fig:fig4}A, SI Movie 2). We quantified performance using wave efficiency $\eta$, defined as forward displacement per cycle normalized by body length, which captures how effectively the prescribed body wave is converted into net translation. Varying $A$ and $\xi$ across the robot's mechanical workspace revealed that swimming efficiency depends on wave geometry rather than simply increasing lateral bending amplitude: the highest $\eta$ emerged at intermediate values of $A = 40^\circ$ and $\xi = 0.75$ (Figure~\ref{fig:fig4}B), and this gait was used for all subsequent experiments unless otherwise stated.

We then varied $\omega$ to characterize how undulation frequency affects swimming speed. Absolute velocity increased with frequency over the undulatory regime, where the robot maintained a coordinated traveling wave with clear phase progression across joints (Figure~\ref{fig:fig4}C-i). At higher $\omega$, the required motor speed exceeded the practical execution limit of the cable-driven body, and the prescribed traveling wave gave way to attenuated, nearly synchronous joint motion (Figure~\ref{fig:fig4}C-ii; see Supplementary Text for detailed discussion). Although this oscillatory regime represents an execution limit rather than a designed gait, the resulting straighter body profile reduces lateral span and can facilitate passage through confined geometries, as demonstrated in Figure~\ref{fig:figSI-Tube} and SI Movie 3. Together, these open-water characterizations define the baseline gait conditions used for all subsequent hydro-cluttered locomotion experiments.

\subsection*{Robophysical study reveals that programmable body compliance determines locomotion across hydro-cluttered regimes}

Hydro-cluttered locomotion emerges from the interaction between an imposed body wave and environment-induced deformation: contacts can dissipate propulsive work, interrupt wave propagation, redirect the body, or generate reaction forces that contribute to forward progression. Since generalized compliance $G$ regulates the body's admissible deformation during environmental interaction, we hypothesized that $G$ could serve as a governing variable for locomotor performance. The coupled fluid-contact dynamics make this regime difficult to model or simulate, and the complexity can grow when obstacles can move, deform, and contact the body at multiple points simultaneously. We thus used a robophysical approach, systematically varying robot mechanics and environmental constraint strength across a series of standardized laboratory models of representative hydro-cluttered regimes (SI Movie 3).

\begin{figure}[t!]
\centering
\includegraphics[width=0.64\textwidth]{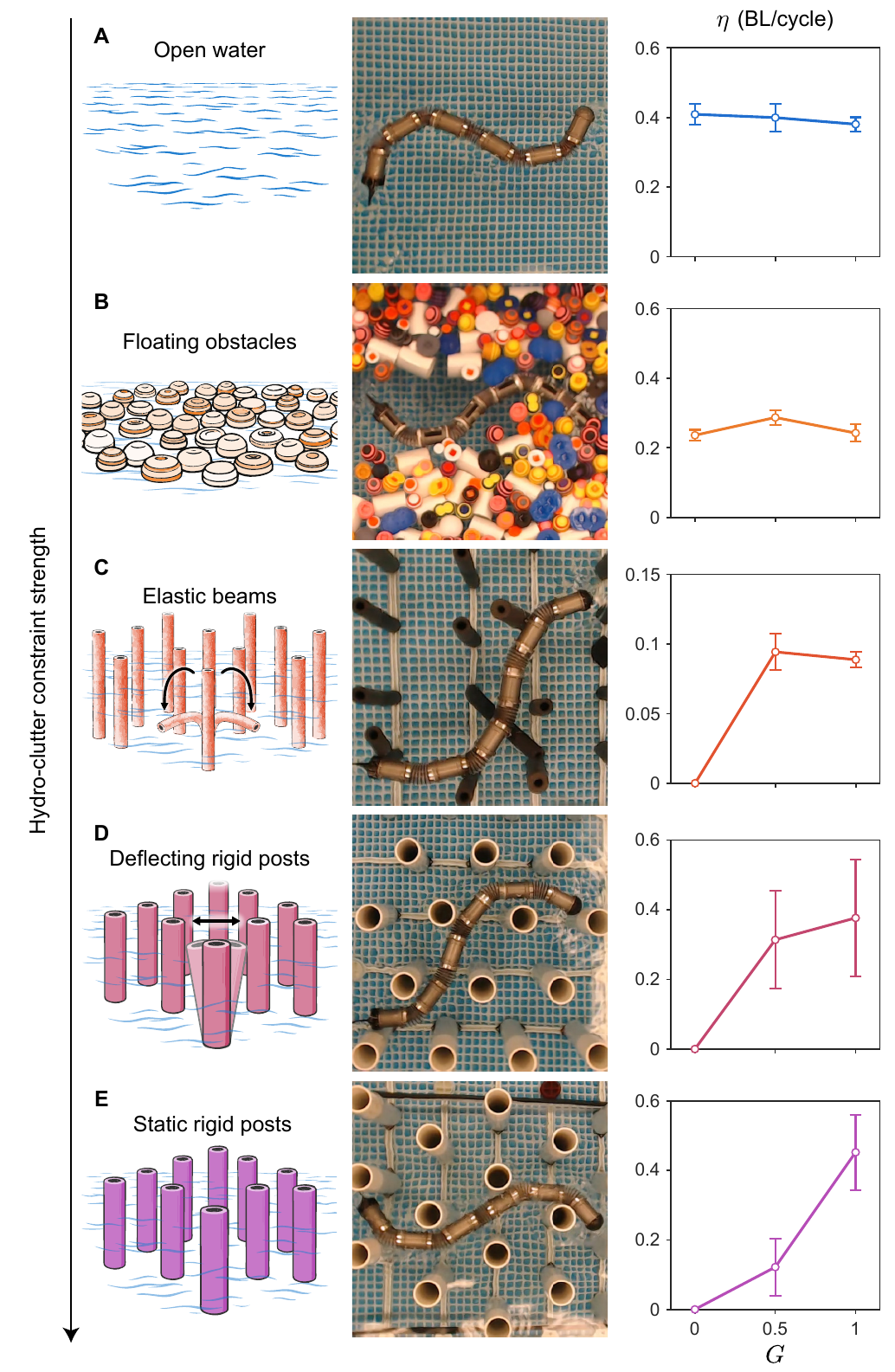} 	
\vspace{-1.5em}
\caption{\textbf{Systematic robophysical experiments show that body compliance facilitates robust locomotion as hydro-clutter constraint strength increases.}
Robophysical experiments comparing AquaMILR locomotion across controlled aquatic environments with increasing hydro-clutter constraint strength: (\textbf{A}) open water, (\textbf{B}) floating obstacles, (\textbf{C}) elastic beams, (\textbf{D}) deflecting rigid posts, and (\textbf{E}) static rigid posts. Each row shows an environment schematic, a representative top-view image, and wave efficiency $\eta$ across generalized compliance $G$. Wave efficiency $\eta$ was measured as displacement per cycle in body lengths. Markers denote means, and error bars indicate SEM across repeated experiments (see also SI Movies 3 and 4).}
\label{fig:fig5}
\end{figure}

We tested this hypothesis across a controlled sequence of five environments with progressively stronger hydro-clutter constraints: open water, floating obstacles, elastic beams, deflecting rigid posts, and static rigid posts (Figure~\ref{fig:fig5}). Floating obstacles could be displaced during contact, elastic beams could deform and recover, deflecting rigid posts could passively sway, and static rigid posts imposed fixed spatial constraints---as constraint strength increased, the mechanical freedom available outside the robot body progressively decreased, and the burden of deformation accommodation shifted increasingly onto the robot body itself. For each environment, we compared three representative compliance states, $G = 0$, $0.5$, and $1$, while holding $\omega = 0.1~\mathrm{Hz}$ to minimize frequency-dependent inertial effects and isolate the role of body compliance in contact-mediated locomotion. Locomotor performance (speed and efficiency) was quantified using wave efficiency $\eta$ (Figure~\ref{fig:fig5}) and mechanical cost of transport $c_{mt}$ (Figure~\ref{fig:figSI-COT}, see Material and Method for detailed calculation method).

In open water (Figure~\ref{fig:fig5}A), varying $G$ produced only modest changes in performance. Wave efficiency decreased slightly with increasing $G$: slacker cables transmit the traveling wave less precisely, reducing forward displacement per cycle. Despite this, $c_{mt}$ also decreased slightly with increasing $G$, since slack cables no longer need to actively resist hydrodynamic forces acting on the body, reducing actuator work by more than the reduction in displacement. Both effects are small, reflecting the dominant requirement in open water, where propulsion is governed by maintaining a coherent traveling body wave. This contrasts with stiffness-tuned fish-like propulsors, where changing stiffness substantially alters the passive dynamics of an oscillating tail or foil and thereby improves open-water speed and efficiency~\cite{zhong2021tunable}. In AquaMILR, $G$ regulates the body's response to environmental interaction rather than optimizing propulsion itself, and this distinction matters least in open water, where contact is absent.

In floating obstacles (Figure~\ref{fig:fig5}B), which represent mobile hydro-clutter such as floating vegetation or suspended debris, the directionally compliant state, $G = 0.5$, produced the highest locomotor performance. Mobile obstacles primarily absorbed propulsive work rather than providing stable reaction forces for forward progression, making this a dissipative clutter regime. At $G = 0$, the non-compliant body could not locally yield to obstacle contact, and the obstacles impeded wave propagation. At $G = 1$, the body was too compliant to transmit the traveling wave effectively through repeated contacts. At $G = 0.5$, directional compliance admitted local contact-induced deformation while preserving sufficient mechanical constraint to transmit the traveling wave, producing the best performance.

In elastic beam arrays (Figure~\ref{fig:fig5}C), which represent rooted flexible structures such as submerged vegetation, the directionally compliant state, $G = 0.5$, again produced the highest locomotor performance. Compared with floating obstacles, this regime imposed stronger constraints since contact was coupled to repeated environmental deformation and recovery. Unlike floating obstacles that are displaced and cleared, elastic beams spring back after contact, repeatedly striking the body and disrupting wave propagation throughout traversal. At $G = 0$, the non-compliant body entangled with the beams, which acted as local barriers to wave propagation and led to traversal failures. At $G = 1$, the body conformed to the beam field but was too compliant to transmit the traveling wave effectively. At $G = 0.5$, directional compliance admitted contact-induced body deformation while preserving sufficient wave transmission to push through the recoverably deformable beam field.

In deflecting rigid posts (Figure~\ref{fig:fig5}D), which represent rigid but movable structures such as swaying stems, roots, or partially constrained debris, bidirectional compliance, $G = 1$, produced the highest locomotor performance. At $G = 0$, the non-compliant body became locally stuck against the posts, as the rigid body could not deform around the posts and the posts could not move out of the way. At $G = 0.5$, directional compliance helped the body respond to contact but limited local shape adjustment to one direction, which was insufficient to navigate the range of contact geometries imposed by the posts. At $G = 1$, each contacted joint could deflect in either direction, allowing the body to passively adapt its local shape around the surrounding posts while continuing to propagate the traveling wave, and contact-induced deformation was distributed across the body. Notably, absolute speed increased substantially compared with elastic beam arrays, and $c_{mt}$ decreased---stronger environmental constraints produced better locomotion. This result indicates that locomotion was no longer dominated by self-generated hydrodynamic propulsion alone. The posts were stiff enough to provide stable reaction forces, and once the body could adapt around them, post contact naturally redirected body deformation into forward progression.

In static rigid posts (Figure~\ref{fig:fig5}E), which represent fixed root networks, rigid vegetation, or submerged structural clutter, the constraints are strongest among all tested regimes. At $G = 0$, the non-compliant body failed as it could not deform around fixed geometric constraints during wave propagation. At $G = 0.5$, the body failed more frequently than in deflecting posts since directional compliance could not accommodate the wider range of contact geometries imposed by the fixed obstacle field. At $G = 1$, the body complied in either direction and recovered its propagating shape after each contact, enabling sustained forward progression. Note that absolute speed at $G = 1$ was higher here than in deflecting rigid posts, and $c_{mt}$ was lower---the strongest tested constraints produced the fastest locomotion. This result sharpens the principle established in the deflecting post regime: the stiffer the posts, the more stable the reaction forces they provide, and the more effectively body compliance can convert contact into ``obstacle-aided" forward progression~\cite{transeth2008snake}. The locomotor principle here is no longer self-generated hydrodynamic propulsion, and the environment itself becomes the primary driver of forward progression, with the body's compliance determining how effectively contact forces are harvested.

These robophysical experiments establish hydro-cluttered locomotion as a regime in which environmental constraints and body-environment interactions dominate locomotor performance. Generalized compliance $G$ serves as a governing variable, with the better-performing state shifting from directional compliance in mobile or recoverably deformable clutter to bidirectional compliance under harsh rigid constraints. As constraints strengthened, successful locomotion depended less on self-generated hydrodynamic propulsion and more on the body's ability to comply with, load against, and recover from environmental contact, giving rise to environment-dominated propulsion. All of these behaviors emerged under the same open-loop undulatory control structure, without sensing, feedback, or environment-specific planning, demonstrating mechanical intelligence at an appropriate level of body compliance (SI Movie 4). The surrounding hydro-cluttered environment is not merely an external disturbance but a non-negligible component of that intelligence: through contact, it filters, redirects, and transmits the robot's body-deformation work, selecting the deformation modes that can be converted into forward progression.

\subsection*{Open-loop depth regulation expands hydro-cluttered locomotion into three dimensions}

\begin{figure}[t!]
\centering
\includegraphics[width=0.8\textwidth]{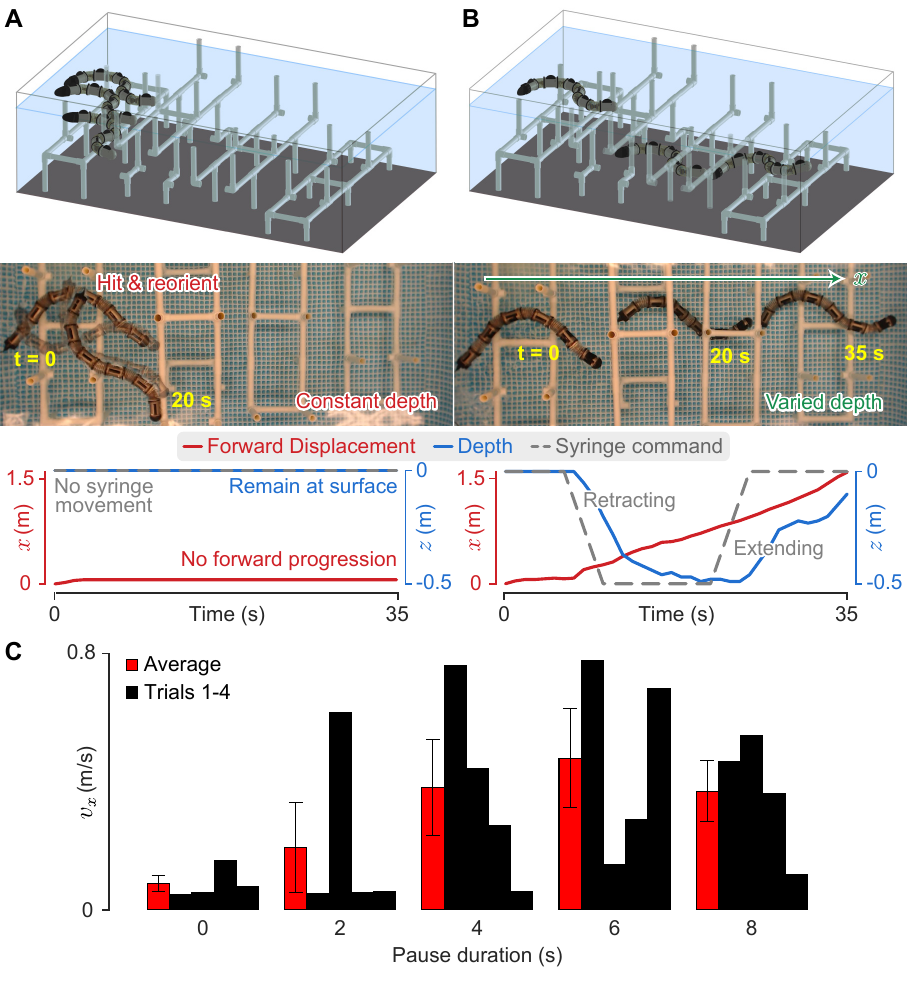} 
\vspace{-1em}
\caption{\textbf{Open-loop depth regulation enables three-dimensional navigation through rigid obstacle networks.}
(\textbf{A}) Representative experiments in a rigid 3D obstacle network. At constant depth, the robot remains near the surface, contacts the structure, and reorients without net forward progression. 
(\textbf{B}) With open-loop depth regulation, the robot alternates between surface-level and submerged interaction regimes and traverses the obstacle network. Time series show forward displacement $x$ (red), depth $z$ (blue), and syringe command (gray) for the constant-depth and varied-depth experiments. 
(\textbf{C}) Forward speed $v_x$ as a function of pause duration during open-loop depth-regulated swimming. Pause duration defines the hold time between syringe retraction and extension commands. Red bars denote average speed, black bars show individual repeated experiments, and error bars indicate SEM (see also SI Movie 5).}
\label{fig:fig6}
\end{figure}

Having established body compliance as a governing variable for hydro-cluttered locomotion in two dimensions, we next asked whether depth regulation could extend this capability into three dimensions. We used a static rigid three-dimensional obstacle network as the test environment (Figure~\ref{fig:fig6}), with $G = 1$ fixed throughout to isolate the effect of depth regulation from compliance.

Without syringe actuation, AquaMILR remained at the surface, repeatedly contacted the horizontal obstacles, and failed to make forward progress (Figure~\ref{fig:fig6}A). With open-loop depth regulation, syringe retraction and extension drove the body downward and upward during continued forward undulation, allowing the robot to bypass horizontal obstacles while body compliance handled negotiation of vertical obstacles (Figure~\ref{fig:fig6}B, SI Movie 5). The same undulatory gait that failed at constant depth traversed the full obstacle network with depth variation. This result reveals a complementary relationship between the two control variables: body compliance operates in the lateral plane, absorbing and redirecting contact with vertical structures, while depth regulation operates in the vertical dimension, shifting the body wave away from horizontal barriers. Together, they partition the three-dimensional obstacle negotiation problem in a way that neither mechanism could solve alone.

Depth regulation proved equally effective in a flexible three-dimensional obstacle network composed of elastic beams (Figure~\ref{fig:figSI-Flex3D}). This environment posed a distinct challenge from the rigid case: rather than blocking specific vertical regions, elastic beams actively interfered with the body by springing back after contact, making entanglement the dominant failure mode under constant-depth swimming. Depth variation addressed this by continuously shifting the robot's vertical position during forward undulation, allowing the body to move through regions of the structure before entanglement could develop. That the same open-loop depth regulation strategy handled both rigid and flexible three-dimensional clutter suggests the principle is not specific to a particular obstacle type, but reflects a more general advantage of vertical mobility in complex three-dimensional environments.

To characterize how depth modulation affected traversal, we varied the pause duration between syringe retraction and extension (Figure~\ref{fig:fig6}C). Very short pauses produced insufficient vertical excursion, keeping the robot near the surface where it continued to contact horizontal obstacles, while very long pauses drove the robot beyond the accessible vertical range of the structure. Intermediate pause durations produced the highest speeds, reflecting a match between vertical excursion and the geometry of the obstacle network. Across all tested pause durations, open-loop depth regulation consistently outperformed constant-depth swimming, demonstrating that a simple periodic syringe command was sufficient to enable three-dimensional traversal without any sensing or feedback.

Body compliance and depth regulation together form a layered open-loop control architecture for three-dimensional hydro-cluttered locomotion. Compliance handles lateral body-environment interactions, absorbing and redirecting contact forces, while depth regulation moves the robot through different vertical regions of the structure, bypassing obstacles that block any single depth level. Depth regulation robustifies hydro-cluttered locomotion by giving the robot additional opportunities to escape local blockage, recover from entanglement, and continue through structures where a fixed-depth path fails. Neither mechanism requires sensing, feedback, or environment-specific planning, yet their combination enabled traversal of three-dimensional obstacle networks that neither mechanism could handle alone, demonstrating that open-loop body mechanics alone are sufficient for three-dimensional locomotion through environmental complexity.

\subsection*{Emergent inertia-driven rolling enables recovery from hard constraints}

Body compliance and depth regulation were characterized at low undulation frequency, where inertial effects are minimal. At higher frequencies, however, inertia becomes a non-negligible factor in body-environment interaction, potentially altering how the body responds to contact. We examined how increasing $\omega$ affected AquaMILR's behavior in rigid posts, where low-frequency non-compliant bodies had previously failed.

\begin{figure}[t!]
\centering
\includegraphics[width=0.66\textwidth]{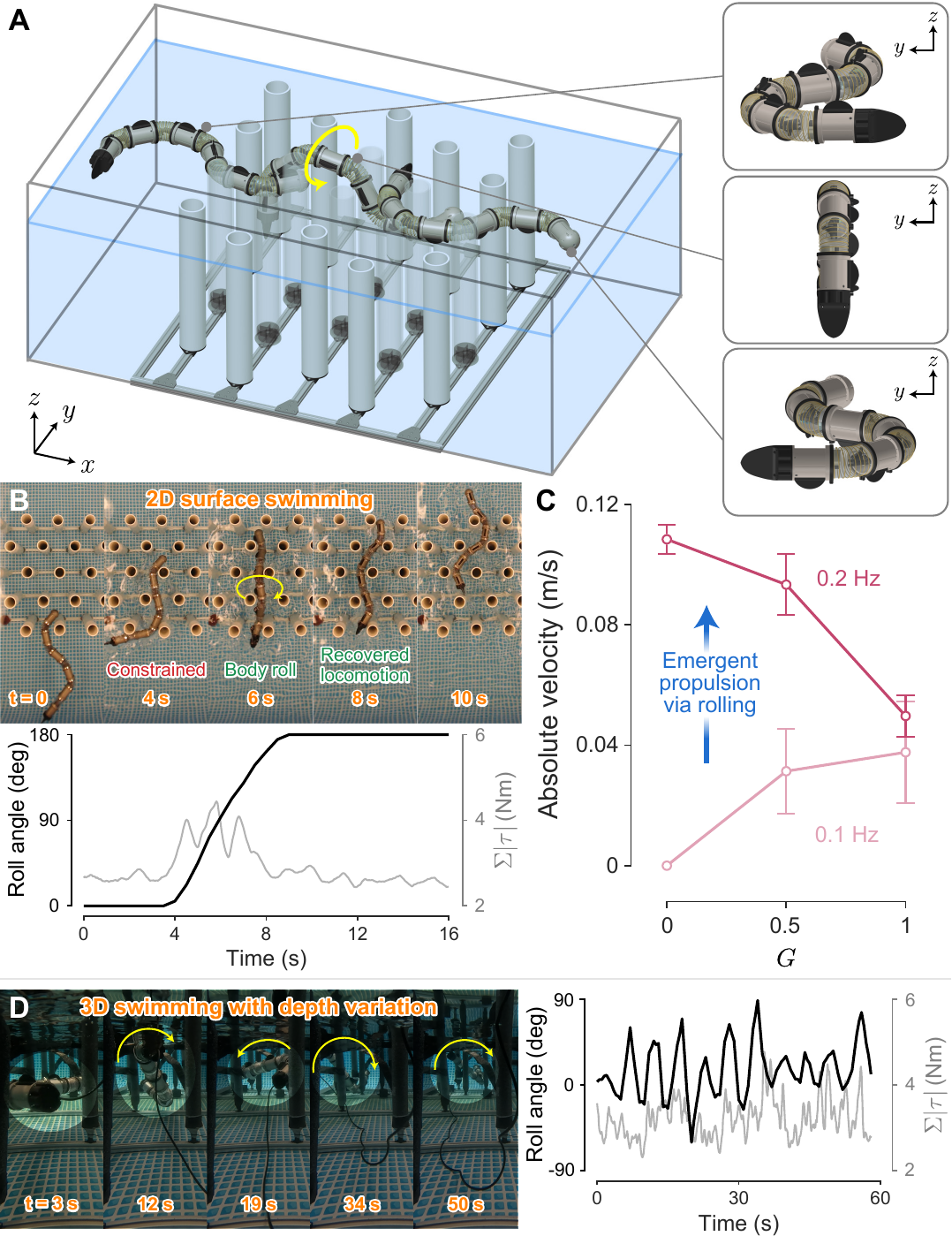} 	
\vspace{-1em}
\caption{\textbf{Emergent body roll creates an alternative locomotor mode under strong environmental constraints.}
(\textbf{A}) Schematic of roll-assisted locomotion in a rigid post array, with representative body postures shown on the right. 
(\textbf{B}) In 2D surface swimming, the robot becomes constrained, undergoes body roll, and recovers planar locomotion. Roll angle is shown in black and summed cable-motor torque magnitude $\Sigma|\tau|$ in gray.
(\textbf{C}) Absolute velocity versus generalized compliance $G$ at low and high undulation frequencies in the rigid post array. High-frequency rolling enhances propulsion in low-compliance bodies. Markers denote means, and error bars indicate SEM across five repeated experiments.
(\textbf{D}) In 3D depth-varying traversal, the robot repeatedly rolls while negotiating an obstacle network. Roll angle and summed torque are shown on the right (see also SI Movie 6).}
\label{fig:fig7}
\end{figure}

In the previous experiments, the non-compliant body, $G = 0$, failed completely in rigid posts at $\omega = 0.1~\mathrm{Hz}$, as the body became locally wedged and could not accommodate fixed geometric constraints during wave propagation. Surprisingly, when the same non-compliant body was driven at an increased undulation frequency, $\omega = 0.2~\mathrm{Hz}$, it did not simply remain jammed. Instead, AquaMILR exhibited emergent body rolling during constrained surface swimming (Figure~\ref{fig:fig7}A and B, SI Movie 6). The robot became locally constrained when several body segments loaded against the posts, and the summed cable-motor torque magnitude increased as the gait continued against the geometric constraint. This buildup of motor torque, combined with contact forces from the posts and inertia from the continued body motion, rotated the robot out of plane. The resulting change in contact geometry allowed the robot to release the planar constraint and resume forward locomotion, showing that hard contact can trigger a new locomotor mode.

Increasing $\omega$ reversed the relationship between compliance and locomotor performance (Figure~\ref{fig:fig7}C, SI Movie 6). At low frequency, low-$G$ bodies remained prone to jamming since they could neither deform sufficiently in plane nor generate enough inertial response to escape, and higher compliance produced better performance. At higher frequency, low-compliance bodies instead achieved greater forward speed through rolling-assisted recovery, outperforming their high-compliance counterparts. Body roll per cycle decreased with increasing $G$, showing that constraint-induced forces were absorbed by local compliance at high $G$, but transmitted into whole-body rotation at low $G$, where inertia became a torque source for rolling than a source of jamming.

Rolling also appeared during depth-regulated swimming through a flexible three-dimensional obstacle network (Figure~\ref{fig:fig7}D). Unlike the two-dimensional case where rolling was a one-time escape event, the robot here repeatedly rolled and rebalanced throughout traversal as the body contacted vertical and horizontal obstacles during depth variation. This recurring out-of-plane motion maintained forward progression through the structure, suggesting that in three-dimensional clutter, rolling functions less as a recovery mechanism and more as a continuous rebalancing mode that the body cycles through during locomotion.

These results reveal how compliance and inertia interact as complementary mechanisms in hydro-cluttered locomotion. At low $G$ and low $\omega$, the body can neither deform sufficiently to accommodate hard constraints nor generate enough inertial response to escape, leading to jamming. At low $G$ and higher $\omega$, inertia compensates for insufficient compliance, converting contact loading into whole-body rotation that resolves the constraint. At high $G$, local compliance absorbs contact loading regardless of frequency, and rolling is suppressed. In terrestrial limbless locomotion, high damping makes inertial effects negligible, and compliance is the only mechanism available for contact negotiation~\cite{wang2023mechanical}. In the aquatic setting, inertia is non-negligible and introduces an additional recovery pathway when compliance alone is insufficient. During depth-regulated three-dimensional traversal, rolling functioned not as a one-time escape but as a recurring rebalancing mode, with the body continuously cycling through different orientations while maintaining forward progression. Further, rolling does not interrupt forward progression in an undulatory limbless robot. The cylindrical body cross-section and the nature of undulatory propulsion enable the body to generate forward thrust across all roll orientations, without losing propulsion upon rolling. Rolling is thus an emergent secondary locomotor mode that exploits the inertial properties, converting otherwise failure-inducing contact into self-organized recovery without sensing, gait switching, or feedback control.

\subsection*{Field deployment demonstrates environmental inspection in hard-to-access aquatic habitats}

\begin{figure}[t!]
\centering
\includegraphics[width=0.85\textwidth]{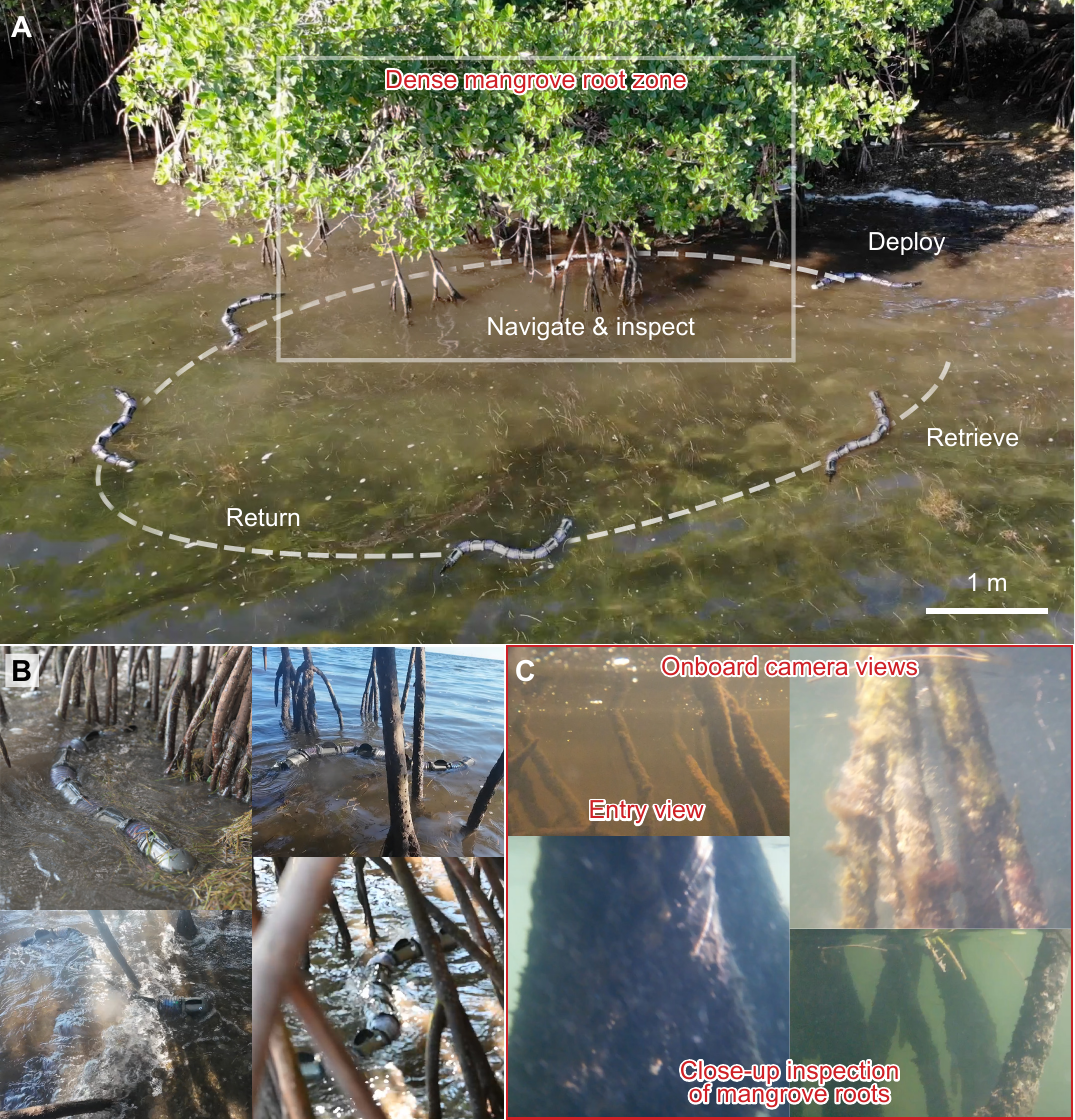} 	
\caption{\textbf{Field deployment demonstrates mangrove root inspection in a hard-to-access aquatic habitat.}
(\textbf{A}) Overview of a mangrove inspection mission. AquaMILR is deployed, navigates in a dense mangrove root zone, inspects the root region, returns, and is retrieved. The dashed curve indicates the approximate robot trajectory, and the scale bar denotes 1 m.
(\textbf{B}) Representative external views of AquaMILR operating near mangrove roots, shallow boundaries, surface vegetation, and cluttered water during the mission. These scenes show contact-rich interactions with natural environmental structures.
(\textbf{C}) Representative onboard camera views acquired during navigation and inspection. The images show the entry view into the root zone and close-up inspection views of submerged mangrove roots (see also SI Movie 7).}
\label{fig:fig8}
\end{figure}

Mangrove root zones present branching rigid roots, surface vegetation, soft substrates, shallow boundaries, and confined passages simultaneously, forming a composite hydro-cluttered environment spanning the full range of conditions studied in the laboratory. Human access to these zones is limited, and they require robotic inspection that conventional underwater platforms struggle to provide. AquaMILR was deployed in a natural mangrove habitat to perform environmental inspection of the root zone, demonstrating whether the locomotor principles established in the robophysical experiments transfer to field operation (Figure~\ref{fig:fig8}, SI Movie 7).

Throughout the mission, AquaMILR operated entirely under open-loop undulatory locomotion with no prior knowledge of the environment and no real-time sensing-based adjustment. A constant offset was added to the nominal gait to bias the turning direction, producing a large arc trajectory that carried the robot from the deployment region into the root zone and back for retrieval (Figure~\ref{fig:fig8}A).

As the robot moved through the habitat, the mechanical capabilities characterized in the robophysical experiments engaged naturally in response to the environment (Figure~\ref{fig:fig8}B). Body compliance negotiated contact with branching roots and confined passages, depth regulation shifted the robot between surface and submerged interactions, and body rolling resolved strong local constraints encountered along the way. None of these responses were planned or triggered by sensing; they emerged from the same open-loop mechanical properties that had been studied in the laboratory.

The onboard camera acquired visual data of the root zone throughout the mission, including close-up views of submerged root structures that are inaccessible to human observers (Figure~\ref{fig:fig8}C). Through mechanical intelligence alone, the robot can already support practical tasks in natural, cluttered, and human-inaccessible aquatic habitats. AquaMILR can preserve limited onboard power, computation, communication bandwidth, and payload capacity for task-level sensing and environmental data collection.

\section*{Discussion}

Hydro-cluttered aquatic environments define a unique locomotor regime in which swimming is inseparable from contact, deformation, confinement, buoyancy, and recovery. Through systematic robophysics studies on AquaMILR, we identified three complementary principles for hydro-cluttered locomotion, each operating in a distinct physical domain. Programmable compliance governed lateral body-environment interactions, determining whether contact-induced deformation was dissipated or redirected into forward progression. Open-loop depth regulation extended this to the vertical dimension, allowing the same undulatory gait to access three-dimensional space through obstacle networks. Emergent inertia-induced rolling added a third layer, where planar compliance was insufficient, inertial loading converted hard contact into whole-body reorientation and recovery. Notably, rolling did not interrupt forward motion, pointing to a broader advantage of undulatory locomotion: the symmetric cylindrical body geometry allows propulsion to persist regardless of orientation, making out-of-plane reorientation part of recovery instead of a terminal failure mode.

These results extend mechanical intelligence~\cite{sitti2021physical,pfeifer2006body,webster2026interdisciplinary,laschi2016lessons} in limbless robots from terrestrial to hydro-cluttered aquatic locomotion. The core idea that programmable body compliance can simplify control carries over, but the aquatic setting introduces complications absent in terrestrial systems, including three-dimensionality, inertia, and strict constraints on onboard power, computation, bandwidth, and payload. Recent work on terrestrial mobile robots already shows that computation and electronics can consume over half the total energy budget~\cite{bjelonic2025towards}, and this burden is likely heavier in hydro-cluttered aquatic systems, where communication constraints and limited payload capacity leave little room for computation-heavy solutions. AquaMILR addresses part of this mechanically, reducing the need to continuously detect and respond to every local contact event. Further, mechanical intelligence here is not a property of the robot alone. Through contact, the hydro-cluttered environment acts as a physical filter and transmitter, selecting and redirecting the body deformation modes that can be converted into forward progression, thus is a non-negligible component of mechanical intelligence. 

The mangrove deployment showed that open-loop mechanical intelligence alone is sufficient for practical environmental inspection, but it also defines a natural path toward higher autonomy. Mechanical intelligence reduces the burden on low-level perception and control, freeing onboard computation for higher-level tasks. Future systems could layer adaptive feedback on top of this mechanical foundation, combining compliance tuning, visual-assisted navigation, or adaptive depth regulation with the passive robustness demonstrated here, without sacrificing the resource efficiency that makes field deployment practical.

AquaMILR suggests a broader design paradigm for hydro-cluttered robotics, in which environmental complexity is not an obstacle to be avoided but a resource to be regulated and exploited through body-environment coupling. Whole-body undulation also offers practical advantages over propeller-based systems such as quieter operation, reduced localized flow disruption, and safer physical interaction with the environment and its inhabitants. These properties create opportunities in settings where conventional robots are limited by collision avoidance, propeller safety, or access constraints, including environmental and infrastructure inspection, habitat and aquaculture monitoring, and search-and-rescue in confined aquatic environments.




\clearpage 

\bibliography{mainbib}
\bibliographystyle{sciencemag}


\section*{Acknowledgments}
The authors would like to thank C. Pierce for proofreading the manuscript; M. Han for data acquisition and analysis; M. Quirke-Shattuck for safety support and coordination during the mangrove field test; D. Cederberg, K. Duwin, E. Williams, L. Green, A. Asghar, and C. White for assistance during the mangrove field test; and the administration of Ransom Everglades School for providing access to the facilities used for the field test.
\paragraph*{Funding:}
This work was supported by Army Research Office Grants No. W911NF-11-1-0514 and No. W911NF20-1-0129.
\paragraph*{Author contributions:}
Conceptualization: T.W., M.F., and D.I.G. Methodology: T.W., M.F., G.T., N.C., J.D., Z.J.X., D.D., S.L., and D.P. Investigation: T.W., M.F., G.T., N.C., J.D., Z.J.X., D.D., S.L., D.P., J.L., and E.G. Funding acquisition: D.I.G. Project administration: T.W., M.F., E.G., and D.I.G. Supervision: D.I.G. Writing, original draft: T.W., M.F., G.T., and P.M. Writing, review \& editing: T.W., M.F., G.T., P.M., and D.I.G.
\paragraph*{Competing interests:}
There are no competing interests to declare.
\paragraph*{Data and materials availability:}
All data needed to evaluate the conclusions in this paper are present in the paper or the Supplementary Materials.


\subsection*{Supplementary materials}
Materials and Methods\\
Supplementary Text\\
Figs. S1 to S7\\
Table S1\\
References \textit{(96-\arabic{enumiv})}\\ 
Movies S1-S7


\newpage


\renewcommand{\thefigure}{S\arabic{figure}}
\renewcommand{\thetable}{S\arabic{table}}
\renewcommand{\theequation}{S\arabic{equation}}
\renewcommand{\thepage}{S\arabic{page}}
\setcounter{figure}{0}
\setcounter{table}{0}
\setcounter{equation}{0}
\setcounter{page}{1} 


\begin{center}
\section*{Supplementary Materials for\\ \scititle}

Tianyu Wang$^{1,2,3,\dagger}$,
Matthew Fernandez$^{3,\dagger}$,
Galen Tunnicliffe$^{1,3}$, \\
Nikolas Cornell$^{3}$,
Justin Duong$^{4}$,
Zhaochen J. Xu$^{2}$,
Patricia Meza$^{5}$,\\
Donoven Dortilus$^{1,3}$,
Sean Lublinsky$^{2}$,
Darsh Parikh$^{4}$,\\
Jianfeng Lin$^{2}$,
Emily Grace$^{6}$,
Daniel I. Goldman$^{1, 2, \ast}$\\
\small$^\ast$Corresponding author. Email: daniel.goldman@physics.gatech.edu\\
\small$^\dagger$These authors contributed equally to this work.
\end{center}

\subsubsection*{This PDF file includes:}
Materials and Methods\\
Supplementary Text\\
Figures S1 to S7\\
Table S1\\
Captions for Movies S1 to S7

\subsubsection*{Other Supplementary Materials for this manuscript:}
Movies S1 to S7

\newpage


\subsection*{Materials and Methods}

\subsubsection*{Robot design and mechanical architecture}

\begin{figure}
\centering
\includegraphics[width=1\textwidth]{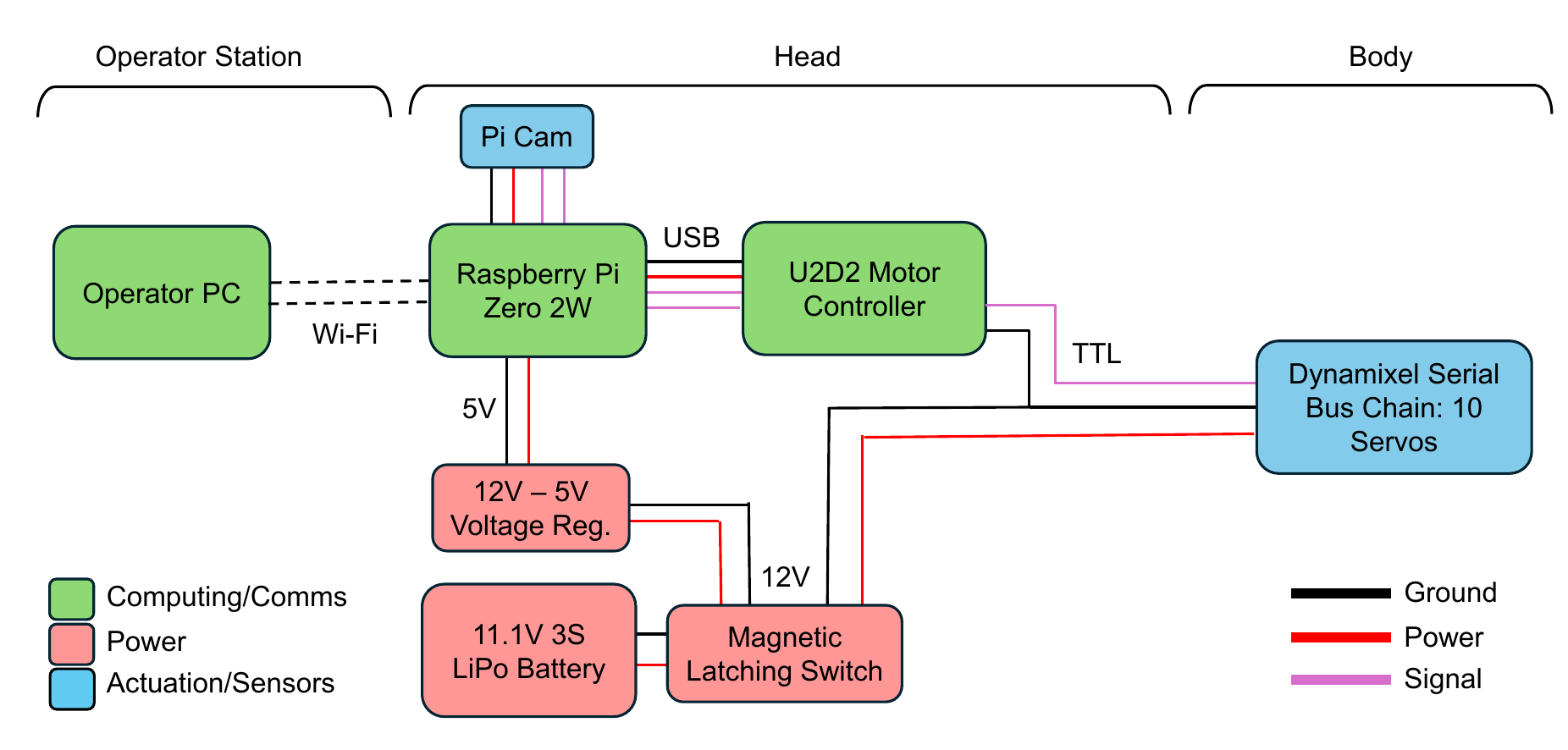}
\caption{\textbf{Electrical architecture of AquaMILR.}
System-level wiring diagram showing the operator-side PC, onboard head electronics, and body actuation network. The operator-side PC communicates with the onboard Raspberry Pi Zero 2W through Wi-Fi. The onboard computer interfaces with the camera, voltage divider, and U2D2 motor controller, which drives the Dynamixel serial bus chain in the body through TTL communication. Power is supplied by an 11.1 V 3S LiPo battery through a magnetic latching switch, with regulated 5 V power for onboard electronics and 12 V power for the motor bus. Component colors indicate computing/communication, power, and actuation/sensing subsystems, and line colors indicate ground, power, and signal connections.}
\label{fig:SI-electronics}
\end{figure}

AquaMILR was constructed as a modular aquatic limbless robot consisting of six serially connected body modules and five actuated joints. Each module is 0.175 m long, giving the assembled robot a total body length of approximately 1.1 m. The head module housed the primary electronics and visual payload, while the remaining modules contained joint actuation and depth-regulation components. The electrical architecture of the robot is provided in Figure~\ref{fig:SI-electronics}.

Each module used a machined 6061 aluminum outer casing with an anodized aluminum oxide coating. The outer casing had an outer diameter of 63 mm and flanged ends with a 75 mm diameter. Each flange contained an O-ring groove and an eight-screw hole pattern, forming a gasket face seal between neighboring components. This flange seal provided the primary waterproofing for the rigid body modules.

Adjacent modules were connected using flexible PVC duct tubing with a 2.5-inch diameter. The tubing formed a puncture-resistant flexible section between rigid modules, allowing bending between neighboring body segments. Each end of the flexible tube was bonded to a flanged aluminum cuff using 3M Adhesive Sealant 5200, and the cuff mated with the neighboring module through the O-ring face seal. 3D-printed tube clamps were used to mechanically secure the tubing to the cuffs. This connection maintained watertight sealing while allowing the robot body to bend during locomotion.

External fins could be attached to vertical and lateral slots on the module casings to provide additional stability and hydrodynamic surface area. For a subset of controlled laboratory experiments, the tail fin was replaced with a tethered tail assembly to facilitate robot retrieval and avoid battery replacement during repeated trials.

\noindent \underline{Cable-driven joint hardware}

\noindent Five body modules contained bilateral cable-drive hardware for actuating the joints between adjacent modules. Each joint used left and right cables routed across the joint and driven by a DYNAMIXEL 2XC430-W250-T dual-axis servo motor. Each cable was wound around a 10 mm diameter pulley mounted to a motor output shaft and anchored on the opposite side of the joint. Non-elastic Rikimaru braided fishing line with a rated maximum load of 800 N was used as the cable material. A clevis-pin joint constrained the main bending degree of freedom between modules. Differential cable motion generated lateral bending at the joint, allowing coordinated joint commands to produce undulatory body motion.

\noindent \underline{Depth-regulation hardware}

\noindent Depth regulation was implemented through syringe extension and retraction within individual body modules. Each depth-regulation unit used a syringe-based mechanism to adjust the mass of the corresponding module. Because the robot was trimmed near neutral buoyancy, coordinated syringe actuation changed its effective density, shifting it between floating, sinking, and near-neutral states during locomotion.

Each depth-control unit contained a pair of custom machined aluminum D-profile syringe chambers packaged beneath the primary cable-drive motor. 3D-printed plungers with perimeter O-rings sealed the syringe chambers and moved linearly during syringe extension and retraction. Plunger motion was driven by a Dynamixel XC330-T288-T servo motor coupled to a drive pulley, which used a belt transmission to control a driven pair of pulleys. These driven pulleys were attached to threaded nuts that interfaced with bushing-supported leadscrews, and each plunger was clamped to a leadscrew to allow linear displacement inside the syringe chamber. A pulley ratio of 15:6 was used to allow an actuation speed of up to 25 mm/s. Water exchange occurred through a sealed barbed fitting connected to silicone tubing and an SLA 3D-printed T-nozzle, which routed flow to the syringe chambers. Each system also included an adjustable-height bearing to tension the drive belt.

\noindent \underline{Onboard electronics, visual payload, and communication}

\noindent The head module housed onboard power, computation, visual inspection, and communication hardware. A Raspberry Pi Zero 2W provided onboard computation and enabled Wi-Fi communication with a surface computer when the robot was near the water surface. Power was supplied by a 500 mAh 3S LiPo battery providing 11.1 V nominally. With an average current consumption of 0.5 A, the system had approximately 1 h of battery life. A step-down voltage regulator supplied 5 V power to the Raspberry Pi, camera, and low-voltage electronics.

The visual payload consisted of a 16 MP Raspberry Pi camera for onboard inspection during field deployment. The Raspberry Pi communicated with a Robotis U2D2 motor controller, which communicated with the distributed actuators through TTL protocol. Communication and power were serially connected to all motors along the robot body. A custom MOSFET latching circuit with a magnetic reed switch allowed the robot to be powered on and off using a momentary magnet swipe without opening the waterproof head module.

\noindent \underline{Robot specifications}

\noindent A summary of robot specifications is provided in Table~\ref{tab:robot_specs}.

\subsubsection*{Robophysical experiments}
We built several different laboratory models to test different regimes of aquatic cluttered terrain within an above-ground swimming pool measuring 3 m in length, 2 m in width, and 0.5 m in depth. These models can be separated into 2 categories regarding assembly: anchored and non-anchored. 

The deflecting rigid posts, static rigid posts, elastic beams, and 3D flexible structure were anchored to a substructure that was constructed with 25.4mm square aluminum T-slot extrusion assembled in a rectangular frame with several cross beams that spanned the effective width of the pool. The cross beam separation distances were customizable to achieve different geometries, while the primary nominal distance used between each cross beam was 220 mm. Vertical members for each structure were fastened to the substructure with 3D-printed PLA adapters. These adapters were fastened to the T-slot extrusion with 1/4-20 bolts and flat T-nuts (Amazon), while being press-fit into the vertical members themselves. These vertical members were made from 3.5-inch OD PVC pipe in the case of the deflecting and static posts, while the elastic beams and structure used 1.625-inch OD foam pipe insulation (Home Depot). The PVC members were sufficiently constrained to the adapters with the press fit, but the pipe insulation was cable-tied to the adapters to prevent them from floating off the internal press fit. The floating obstacles used various 3D-printed PLA sphere sections not exceeding 3-inch diameter as well as cut sections of 3-inch OD PVC pipe to simulate surface debris (Figure ~\ref{fig:fig5}). The deflecting rigid posts and static rigid posts both used a triangular lattice arrangement to form a dense distribution (250 mm, or 1/4 BL). The only difference between the deflecting posts and static posts was the presence of an acrylic sheet above the waterline with waterjet holes to hold each post in position. This provided a secondary binding point so the posts could not swaying from the lattice positioning when the robot was interacting. The rigid 3D obstacle network was assembled with 1-inch PVC pipe and connectors (Home Depot). The flexible 3D obstacle network was assembled with elastic beams (pool noodles, 2-inch OD). In current tests, water steam was created by the four thrusters at the opposing end (DOMICA 90 GPH Mini Submersible Pump) (Figure ~\ref{fig:figSI-Stream}). The narrow tube test was a 6-inch tube of acrylic (Figure ~\ref{fig:figSI-Tube}).

A high-resolution camera (Logitech HD Pro Webcam C920) was mounted above the pool to record videos of each experiment for data analysis. 

\subsubsection*{Mechanical cost of transport calculation}
We quantified the energetic cost of locomotion using the mechanical cost of transport, $c_{mt}$, in addition to the kinematic performance metric $\eta$. Whereas $\eta$ measures how much forward motion is produced by each prescribed body wave, $c_{mt}$ estimates how much actuator work is required to produce that motion. This distinction is important in hydro-cluttered environments because environmental contact can either dissipate body-deformation work or redirect it into forward progression. The dimensionless cost of transport has been widely used to compare locomotor economy across animals and robots~\cite{gabrielli1950price,collins2005efficient,seok2013design,saranli2001rhex}.

For each trial, we calculated
\begin{equation}
c_{mt}=\frac{W_{\mathrm{cable}}}{mgd},
\end{equation}
where $W_{\mathrm{cable}}$ is the total mechanical work delivered through the cable-driven joints, $m$ is the robot mass, $g$ is gravitational acceleration, and $d$ is the distance traveled by the robot during the trial. The traveled distance was obtained from video tracking by measuring the robot center of geometry over time and summing the frame-to-frame displacement increments.

Cable work was estimated from motor torque and motor encoder measurements. Each cable was driven by a motor-pulley unit, so the cable tension could be inferred from the torque acting about the pulley. However, the raw motor torque includes not only load from the cable, but also the torque required to rotate the motor and pulley through the commanded trajectory. To remove this trajectory-dependent baseline, we first ran a calibration trial for each motor in which the same motor command was executed with the cable detached from the pulley. This no-load measurement provided the baseline torque trajectory $\tau_0(t)$.

During locomotion experiments, the motor torque $\tau(t)$ was recorded at $\Delta t = 10~\mathrm{ms}$ using the internal torque estimate from the servo motor. After subtracting the no-load baseline, the cable tension was estimated as
\begin{equation}
T(t)=\frac{\tau(t)-\tau_0(t)}{R_{\mathrm{pulley}}},
\end{equation}
where $R_{\mathrm{pulley}}$ is the pulley radius. The corresponding cable displacement over one sampling interval was computed from the encoder-measured change in motor shaft angle,
\begin{equation}
\Delta l(t)=R_{\mathrm{pulley}}\Delta \zeta(t),
\end{equation}
where $\Delta \zeta(t)$ is the angular displacement of the pulley during that interval.

The mechanical work associated with one cable was then computed by accumulating the tension-displacement product over the trial,
\begin{equation}
W_j=\sum_t T_j(t)\left|\Delta l_j(t)\right|.
\end{equation}
We used the magnitude of cable displacement so that work input from active cable pulling was accumulated over a gait cycle. The total cable work was obtained by summing across all actuated cables,
\begin{equation}
W_{\mathrm{cable}}=\sum_j W_j .
\end{equation}
This value was then normalized by $mgd$ to obtain the mechanical cost of transport for that trial.


\subsection*{Supplementary Text}

\subsubsection*{Cable length geometry model}

\begin{figure}
\centering
\includegraphics[width=0.65\textwidth]{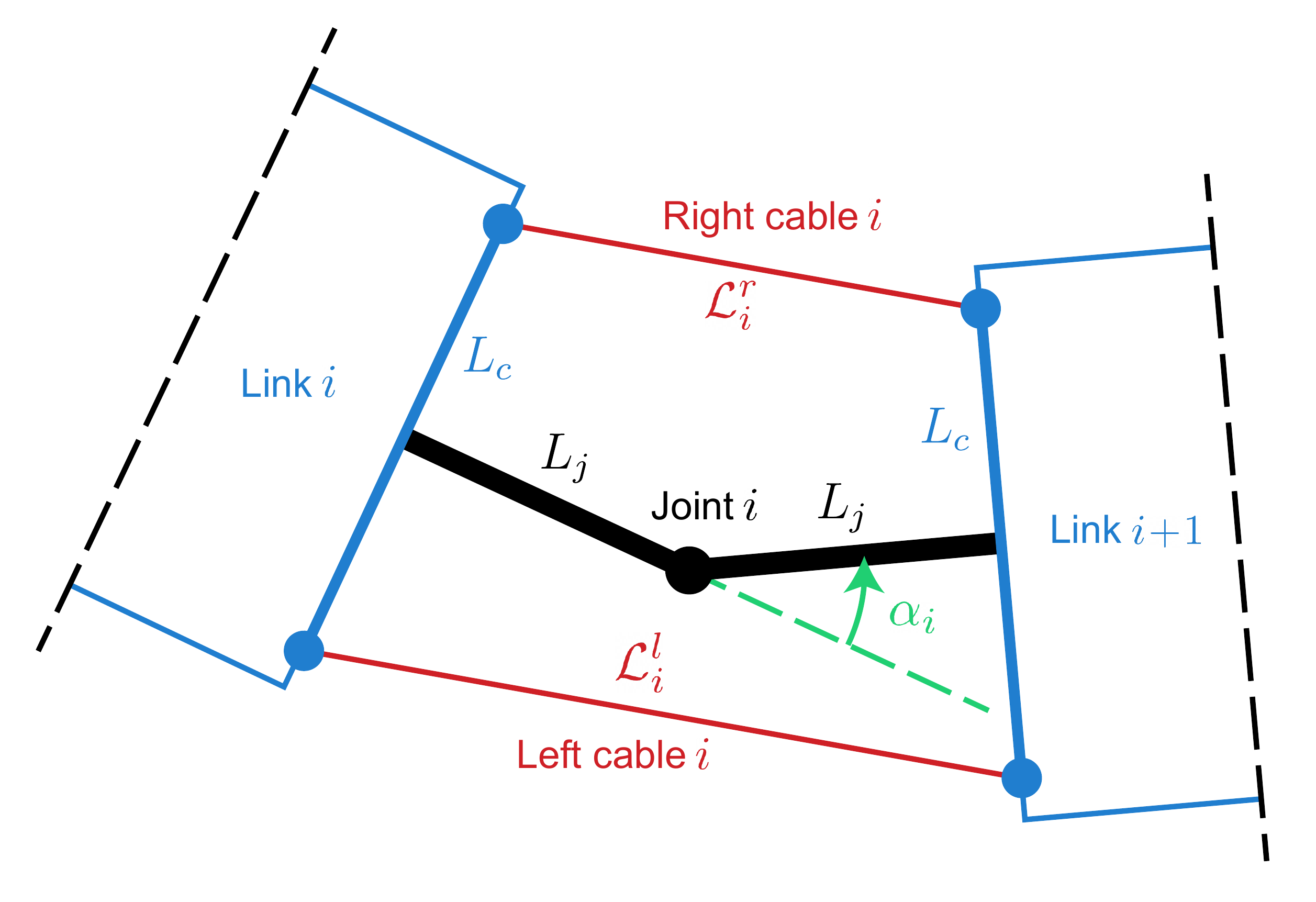}
\caption{\textbf{Geometry of the cable-driven joint used for cable-length calculation.}
Schematic of one bilateral cable-driven joint connecting links $i$ and $i+1$. For a prescribed joint angle $\alpha_i$, the left and right cables have exact taut lengths $\mathcal{L}^{l}_{i}$ and $\mathcal{L}^{r}_{i}$, determined by the joint geometry. The parameters $L_j$ and $L_c$ define the longitudinal joint offset and lateral cable offset, respectively, and are used to convert the desired joint angle into cable length commands.}
\label{fig:figSI-JointGeometry}
\end{figure}

AquaMILR controls each lateral joint through an antagonistic pair of cables routed on the two sides of the joint. To convert a desired joint angle into cable commands, we first defined the cable length that would exactly constrain the joint to a prescribed angle (Figure~\ref{fig:figSI-JointGeometry}). For the $i$-th joint, this prescribed angle is denoted by $\alpha_i$, and the corresponding taut cable lengths on the left and right sides are denoted by $\mathcal{L}^{l}_i(\alpha_i)$ and $\mathcal{L}^{r}_i(\alpha_i)$, respectively. These quantities represent the geometric cable lengths required when the cable is fully engaged, with no intentional slack.

The cable path across a joint can be written from the joint geometry as
\begin{equation}
\begin{aligned}
    \mathcal{L}^{l}_i(\alpha_i) &= 2\sqrt{L_{c}^2 + L_{j}^2}
    \cos\left[-\frac{\alpha_i}{2}+\tan^{-1}\left(\frac{L_{c}}{L_{j}}\right)\right],\\
    \mathcal{L}^{r}_i(\alpha_i) &= 2\sqrt{L_{c}^2 + L_{j}^2}
    \cos\left[\frac{\alpha_i}{2}+\tan^{-1}\left(\frac{L_{c}}{L_{j}}\right)\right],
\end{aligned}
\label{eq:ExactLength}
\end{equation}
where $L_c$ is the lateral cable offset and $L_j$ is the longitudinal joint offset defined by the mechanical layout. Equation~\ref{eq:ExactLength} provides the reference cable lengths for a purely geometrically constrained joint. A non-compliant command uses these exact lengths directly, whereas a compliant command is generated by adding controlled slack to one or both cables.

The generalized compliance parameter $G$ modifies the joint constraint by changing how the commanded cable lengths depart from the exact geometric lengths. At $G=0$, both cables remain taut relative to the desired angle, so the joint is mechanically constrained to follow the prescribed body wave. At $G=0.5$, only the cable opposing curvature-increasing deformation remains engaged, allowing the joint to yield preferentially in one direction. At $G=1$, both sides are relaxed relative to the exact constraint, allowing passive deflection in either direction. Thus, $G$ does not change the prescribed gait itself. Instead, it changes the admissible mechanical response of the joint when fluid forcing or environmental contact perturbs the body.

Cable length commands were implemented through motor-driven pulleys. If $L$ denotes the commanded cable displacement, the corresponding motor position command was calculated as
\begin{equation}
    P(L)=P_0-\kappa L,
\label{eq:MotorPositionCableLength}
\end{equation}
where $P_0$ is the calibrated zero-displacement motor position for that cable, and $\kappa$ converts cable displacement into motor encoder position. Each Dynamixel servo motor provides 4096 encoder positions per full revolution. For a pulley of diameter $D_\mathrm{pulley}$,
\begin{equation}
    \kappa=\frac{4096}{\pi D_\mathrm{pulley}}.
\label{eq:CableMotorConversion}
\end{equation}

The amount of intentional slack introduced by the compliance rule is set by the parameter $l_0$, which converts angular command into additional cable length. In this study, we used $l_0=0.6~\mathrm{mm/deg}$. This value provided sufficient slack to create measurable passive joint freedom while maintaining reliable cable engagement during repeated swimming and obstacle-interaction experiments.

\subsubsection*{High frequency attenuation of prescribed body waves}

In the main text, increasing temporal frequency caused AquaMILR to shift from coordinated traveling wave undulation to higher frequency body oscillation. This oscillatory motion was not prescribed as a separate gait. Instead, it arose when the cable-driven body could no longer fully execute the commanded traveling wave at high undulation frequency.

To examine this transition, we considered how a desired gait is implemented on the robot. For a prescribed body wave with amplitude $A$, spatial frequency $\xi$, and temporal frequency $\omega$, the desired joint angle trajectories are sampled in time and converted into motor position commands. For each gait, this conversion determines the maximum motor speed required to execute the commanded motion. As $A$ or $\omega$ increases, the required motor speed increases, creating practical limits on which prescribed waves can be realized by the robot.

\begin{figure}
\centering
\includegraphics[width=0.5\textwidth]{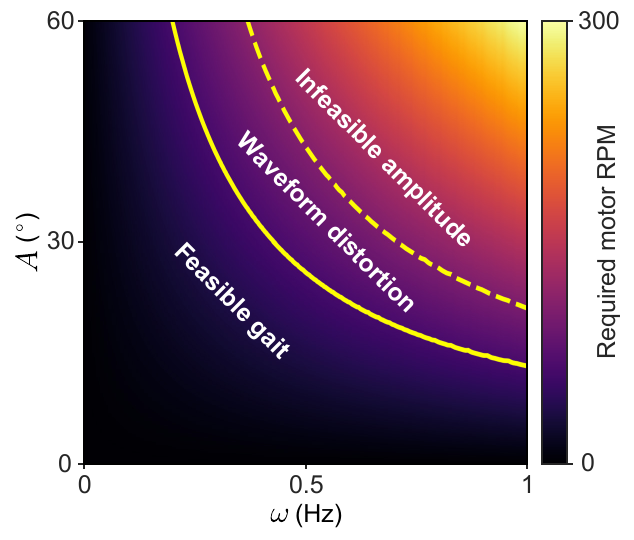}
\caption{\textbf{High frequency attenuation of prescribed body waves.} Estimated gait execution regimes for AquaMILR as a function of gait amplitude $A$ and temporal frequency $\omega$. Color denotes the maximum motor speed required to execute the discretized joint commands. The solid curve marks the boundary beyond which the prescribed traveling wave begins to distort, and the dashed curve marks the boundary beyond which the commanded amplitude can no longer be achieved. These boundaries define three practical execution regimes for this robot: feasible gait, waveform distortion, and infeasible amplitude.}
\label{fig:figSI-GaitRange}
\end{figure}

\begin{figure}
\centering
\includegraphics[width=0.45\textwidth]{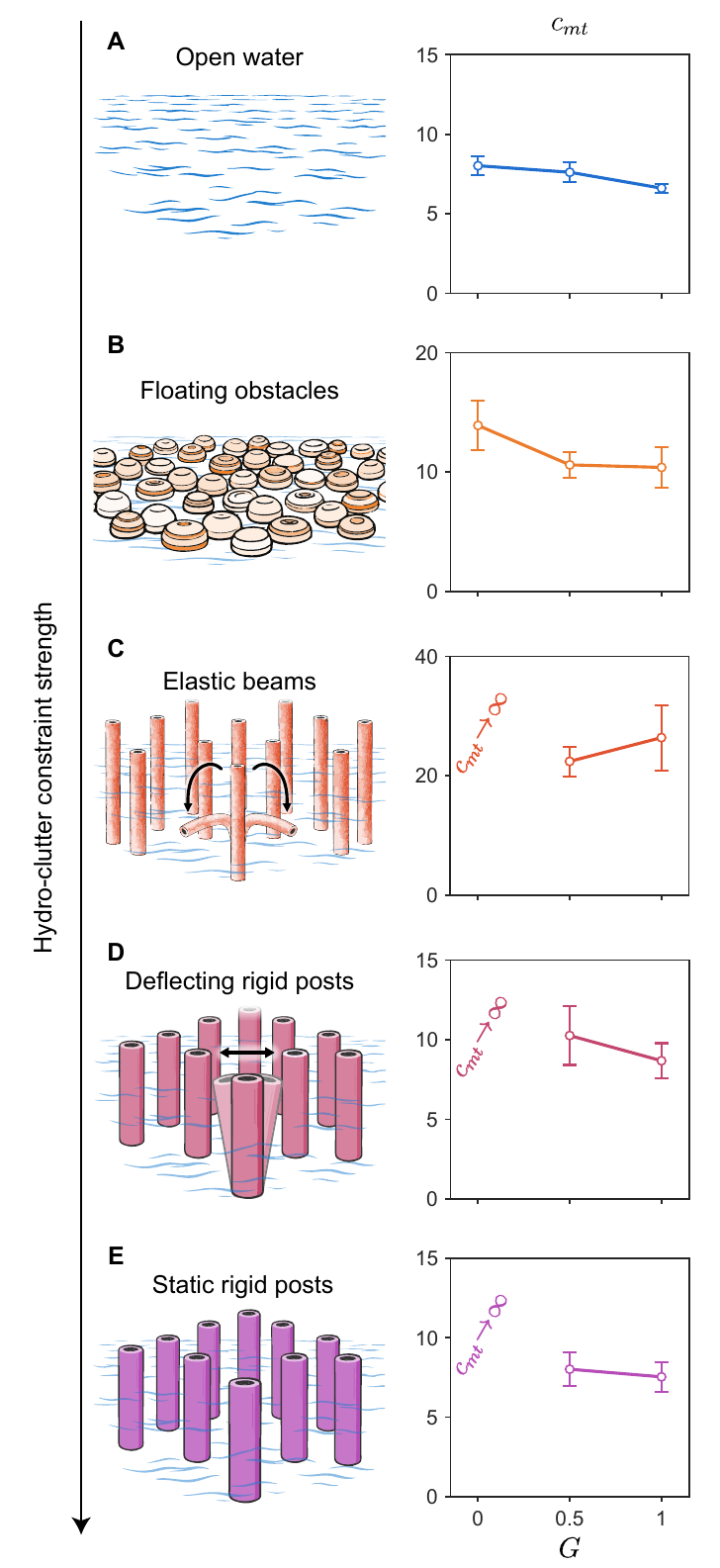}
\caption{\textbf{Mechanical cost of transport across hydro-cluttered environments with increasing constraint strength.}
Robophysical experiments comparing the mechanical cost of transport $c_{mt}$ of AquaMILR locomotion across controlled aquatic environments with increasing hydro-clutter constraint strength: (\textbf{A}) open water, (\textbf{B}) floating obstacles, (\textbf{C}) elastic beams, (\textbf{D}) deflecting rigid posts, and (\textbf{E}) static rigid posts. Each row shows an environment schematic and the corresponding mechanical cost of transport across generalized compliance $G$. Markers denote means, and error bars indicate SEM across repeated experiments.}
\label{fig:figSI-COT}
\end{figure}

This analysis identifies three execution regimes for AquaMILR (Figure~\ref{fig:figSI-GaitRange}). At low required motor speeds, the robot operates in a feasible gait regime, where the commanded traveling wave can be executed reliably. In this regime, the body exhibits coordinated lateral undulation with clear phase progression along the joints. At higher required motor speeds, the robot enters a waveform distortion regime. The motors still drive substantial body motion, but the realized joint trajectories deviate from the prescribed traveling wave, producing distorted phase relationships and reduced wave fidelity. At still higher required motor speeds, the robot enters an infeasible amplitude regime, where the commanded amplitude can no longer be achieved. The realized motion becomes attenuated, with reduced body bending and increasingly synchronous joint motion.

This interpretation is consistent with the joint angle trajectories in Figure~\ref{fig:fig4}C. At low temporal frequency, the measured joint trajectories retain the phase-shifted structure expected for a traveling wave. At high temporal frequency, the trajectories show reduced amplitude and a loss of clear phase progression, indicating that the robot no longer achieves the prescribed body wave and instead oscillates with a straighter body.

\begin{figure}
\centering
\includegraphics[width=0.8\textwidth]{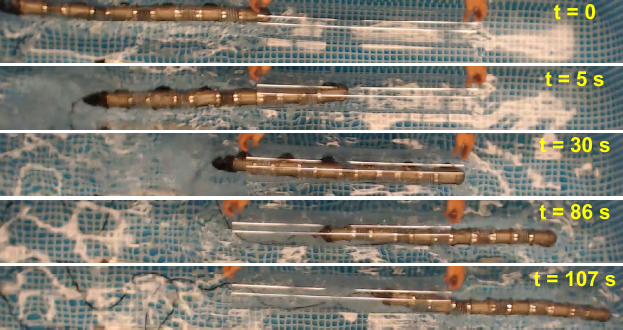}
\caption{\textbf{High frequency oscillation facilitates traversal through confined geometry.}
Sequential frames showing AquaMILR passing through a narrow tube under high-frequency oscillatory body motion. In this regime, the robot does not fully achieve the prescribed traveling wave, but the attenuated, nearly straight body motion reduces lateral excursion and allows the body to move through the confined passage. See also SI Movie 3.}
\label{fig:figSI-Tube}
\end{figure}

\begin{figure}
\centering
\includegraphics[width=0.8\textwidth]{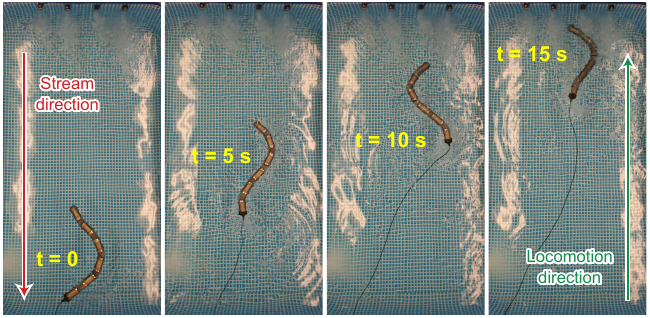}
\caption{\textbf{AquaMILR swims upstream under external flow.}
Image sequence showing AquaMILR locomoting against an imposed surface flow.}
\label{fig:figSI-Stream}
\end{figure}

\begin{figure}
\centering
\includegraphics[width=0.8\textwidth]{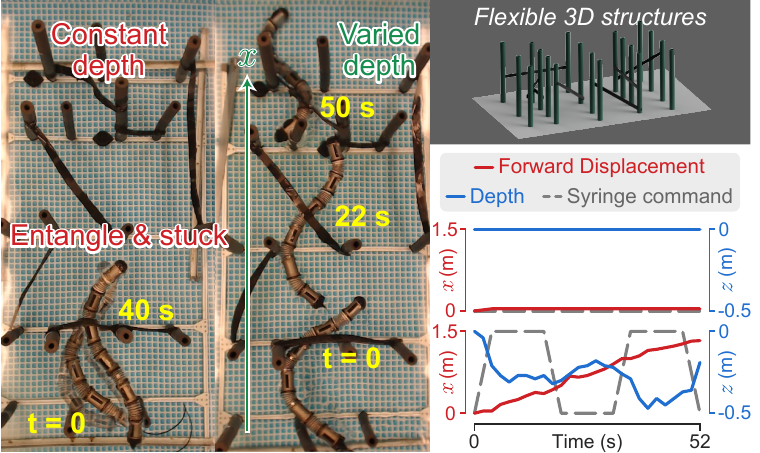}
\caption{\textbf{Depth variation improves locomotion through flexible three-dimensional clutter.}
Representative experiments comparing constant-depth swimming and open-loop depth-regulated swimming in a flexible 3D obstacle network. Under constant-depth swimming, AquaMILR remains near the surface, becomes entangled with the flexible structures, and fails to make forward progress. With varied depth, syringe-driven buoyancy actuation changes the robot's vertical position during undulatory locomotion, allowing the robot to access different regions of the obstacle network and traverse the structure. The schematic shows the flexible 3D obstacle field. Time-series plots show forward displacement $x$, depth $z$, and syringe command for the two representative trials.}
\label{fig:figSI-Flex3D}
\end{figure}

Although this high-frequency oscillatory regime reflects a practical execution limit of the nominal undulatory gait, it can still be functionally useful. Reduced lateral excursion produces a straighter body profile, which can facilitate movement through confined geometries. Figure~\ref{fig:figSI-Tube} and SI Movie 3 show that under high-frequency oscillatory motion, AquaMILR can pass through a narrow tube. This behavior should then be interpreted as a robot-specific execution regime that emerges from the interaction between gait command, actuator limits, and body mechanics, rather than as a separate prescribed gait.



\begin{table}
\centering
\renewcommand{\arraystretch}{1.15}
\begin{tabular}{l l}
\hline\hline
\\[-2em]
\textbf{Specification} & \textbf{Value} \\
\hline\hline
\textbf{Mass} & 
\begin{tabular}[c]{@{}l@{}}
Single module: 0.55 kg \\
Full robot: 3.56 kg
\end{tabular} \\
\hline
\textbf{Dimensions} & 
\begin{tabular}[c]{@{}l@{}}
63 mm body diameter \\
1.1 m total body length
\end{tabular} \\
\hline
\textbf{Power} & 
\begin{tabular}[c]{@{}l@{}}
11.1 V, 0.5 A during open-water operation \\
2 A average during obstacle navigation
\end{tabular} \\
\hline
\textbf{Communication} & 
\begin{tabular}[c]{@{}l@{}}
Internal: TTL communication protocol\\
External: Wi-Fi/Bluetooth communication (Untethered)
\end{tabular} \\
\hline
\textbf{Operational Depth} & $>$3 m (maximum tested depth) \\
\hline
\textbf{Visual payload} & 16 MP Raspberry Pi camera \\
\hline
\textbf{Actuation} & 
\begin{tabular}[c]{@{}l@{}}
Cable motors (DYNAMIXEL 2XC430-W250-T): 1.8 Nm stall torque \\
Syringe motors (DYNAMIXEL XC330-T288-T): 0.92 Nm stall torque
\end{tabular} \\
\hline\hline
\\[-1.25em]
\end{tabular}
\caption{AquaMILR specifications and ratings.}
\label{tab:robot_specs}
\end{table}


\clearpage 



\paragraph{Caption for Movie S1.}
\textbf{Distributed depth-control system enables depth regulation.}
This movie shows AquaMILR regulating its depth in water through coordinated actuation of the distributed depth-control system. By drawing water into or expelling water from the syringe modules, the robot changes its effective buoyancy and transitions among sinking, rising, and near-neutral states. A pool test demonstrates vertical motion driven by syringe actuation alone, without body undulation. A deeper-pool trial demonstrates that the depth-regulation system continues to operate under increased hydrostatic pressure, confirming that the robot remains operational to at least 3 m depth.

\paragraph{Caption for Movie S2.}
\textbf{Gait shape and frequency regulate open-water body motion.}
This movie compares AquaMILR swimming in open water using the baseline gait, a high-spatial-frequency gait, and a high-temporal-frequency gait. The baseline gait uses a serpenoid gait with an amplitude of $A = 40^\circ$, a spatial frequency of $\xi = 0.75$, and a temporal frequency of $\omega = 0.2~\mathrm{Hz}$. The comparison shows how changing the prescribed body wave alters lateral body bending, wave propagation, and forward swimming. Increasing spatial frequency increases the number of body waves along the robot, whereas increasing temporal frequency drives faster body motion and can push the cable-driven body toward an execution limit in which the prescribed traveling wave is attenuated and becomes more oscillatory.

\paragraph{Caption for Movie S3.}
\textbf{Robophysical experiment overview of AquaMILR locomotion across hydro-cluttered regimes.}
This movie summarizes the robophysical study of AquaMILR across controlled laboratory hydro-cluttered aquatic environments. The tested regimes include open water, floating obstacles, elastic beams, deflecting rigid posts, static rigid posts, a narrow tube, and an opposing-current condition. Together, these testbeds span a progression from open-water swimming to contact-rich locomotion involving mobile, compliant, rigid, confined, and flow-disturbed surroundings. These experiments demonstrate that the undulatory body can continue to generate propulsion when environmental forcing is non-negligible.

\paragraph{Caption for Movie S4.}
\textbf{Generalized compliance $G$ governs locomotion across hydro-cluttered regimes.}
This movie compares AquaMILR locomotion across representative generalized compliance values in floating obstacles, elastic beams, deflecting rigid posts, and static rigid posts. By varying $G$ while keeping the commanded gait fixed, these trials show how body compliance changes the robot's admissible mechanical response to environmental contact. In mobile or recoverably deformable clutter, directional compliance, $G = 0.5$, balances wave transmission with local contact-induced deformation. In rigid post regimes, where environmental structures provide less mechanical freedom, bidirectional compliance, $G = 1$, allows the body to conform to stronger geometric constraints, distribute contact-induced deformation, and recover forward progression. These comparisons show that the better-performing compliance state shifts with hydro-clutter constraint strength, revealing how AquaMILR regulates body-environment coupling through programmable mechanical response.

\paragraph{Caption for Movie S5.}
\textbf{Open-loop depth regulation expands hydro-cluttered locomotion into three dimensions.}
This movie shows AquaMILR using integrated open-loop depth regulation to traverse three-dimensional hydro-cluttered environments. Without syringe actuation, the robot remains near the surface, repeatedly contacts the obstacle network, and reorients without net forward progression. With coordinated syringe retraction and extension, the robot changes its effective buoyancy and moves through different depths while maintaining undulatory locomotion. This depth variation shifts the body wave across different vertical regions of the obstacle field, allowing AquaMILR to access three-dimensional space and traverse the environment. The movie demonstrates that depth regulation expands the robot's accessible locomotor space and complements compliance-mediated obstacle negotiation in three-dimensional hydro-cluttered environments.

\paragraph{Caption for Movie S6.}
\textbf{Emergent inertia-driven body rolling enables recovery under strong environmental constraints.}
This movie shows body rolling emerging during AquaMILR locomotion in two-dimensional and three-dimensional rigid lattice environments. Under strong local constraints, the robot can become wedged or redirected by surrounding posts while the open-loop body wave continues to load the environment. This contact loading can trigger out-of-plane body rotation, changing the local contact geometry and allowing the robot to release the planar constraint and recover forward locomotion. In the two-dimensional lattice, rolling acts as an escape-and-recovery event after strong post contact. In the three-dimensional lattice, repeated rolling and rebalancing occur during traversal, showing that inertia can convert otherwise failure-inducing contact into a self-organized recovery mode under strong hydro-cluttered constraints.

\paragraph{Caption for Movie S7.}
\textbf{Field deployment demonstrates navigation and inspection in a mangrove habitat.}
This movie shows AquaMILR operating in a natural mangrove environment to demonstrate navigation and visual inspection in a hard-to-access aquatic habitat. The robot moves through shallow, cluttered water near mangrove roots using open-loop undulatory locomotion and records onboard video of root structures that are difficult for humans to access directly. These field interactions include contact with roots, surface vegetation, soft substrates, boundaries, and confined passages, illustrating how the mechanically mediated locomotor capabilities characterized in the robophysical experiments can support environmental inspection. The movie also includes a shallow-water amphibious attempt under wave disturbance, highlighting additional challenges introduced by natural flow, substrate contact, vegetation, and mixed water-boundary interactions during field deployment.



\end{document}